\documentclass[a4paper,fleqn]{cas-dc}

\usepackage[numbers]{natbib}
\usepackage{subfigure}
\usepackage{longtable}
\usepackage{booktabs}
\def\tsc#1{\csdef{#1}{\textsc{\lowercase{#1}}\xspace}}
\tsc{WGM}
\tsc{QE}
\tsc{EP}
\tsc{PMS}
\tsc{BEC}
\tsc{DE}

\begin{document}
\let\WriteBookmarks\relax
\def\floatpagepagefraction{1}
\def\textpagefraction{.001}

\shorttitle{Adversarial Zero-Shot IoT Anomaly Detection}

\shortauthors{Rezakhani et~al.}

\title [mode = title]{ An Adversarial Zero-Shot Learning Approach for Anomaly Detection in Multivariate IoT Traffic Data}                      
\tnotemark[1]

\tnotetext[1]{{This material is based upon work supported by the National Science Foundation under Grant Numbers   CNS- 2318726, and CNS-2232048.} }


%

\affiliation[1]{organization={Holcombe Department of Electrical and Computer Engineering, Clemson University},
    city={Clemson},
    postcode={29634 SC}, 
    country={USA}}
\author[1]{Mahshid Rezakhani}[type=editor]
\fnmark[1]
\ead{mrezakh@clemson.edu}

\author[1]{Tolunay Seyfi}[type=editor]
\fnmark[1] \cormark[1]
\ead{tseyfi@clemson.edu}

\author[1]{Fatemeh Afghah}[type=editor]
\ead{fafghah@clemson.edu}

\fntext[fn1]{These authors contributed equally to this work.}
\cortext[cor1]{Corresponding author.}

\begin{abstract}
Anomaly detection in Internet of Things (IoT) networks presents unique challenges due to the diversity of devices, lack of labeled data, and domain variability across environments. In this paper, we propose a novel framework for multivariate time-series anomaly detection that leverages adversarial learning and contrastive loss within a sequence-based Variational Autoencoder (VAE) architecture. Our method enables zero-shot domain adaptation by jointly optimizing domain-invariant latent representations and semantically structured embedding spaces, without requiring labeled data or raw feature transfer. To address the heterogeneity of IoT deployments, we introduce encoder and decoder adaptor layers that align feature distributions across domains while preserving contextual semantics. Additionally, we propose a destination-based segmentation strategy to better model real-world communication structures in IoT traffic. Our framework is comprehensively evaluated on six distinct datasets spanning industrial, enterprise, general-purpose, smart home, and military automation domains across 44 transfer scenarios. Experimental results demonstrate strong zero-shot generalization in several
cross-domain settings and competitive performance against a contrastive
domain-adaptation baseline under realistic, heterogeneous, and
privacy-constrained IoT conditions.
\end{abstract}




\maketitle

\section{Introduction}

The rapid proliferation of Internet of Things (IoT) devices across industrial, urban, and defense infrastructures has heightened the risk of operational failures and security breaches. Anomaly detection (AD) in IoT networks is therefore a critical capability: it enables the timely identification of hardware malfunctions, communication disruptions, and malicious cyber-attacks by recognizing deviations from expected traffic patterns \cite{HAIDER2025111484,RAHMAN2025100082}. From a security perspective, anomalies map directly to common network attacks such as scanning, denial-of-service (DoS), botnet propagation, and brute-force intrusions. From a reliability perspective, they also capture system-level degradations such as abnormal packet loss or irregular latency surges that signal hardware or connectivity failures \cite{HAIDER2025111484}.

A large fraction of these anomalies can be inferred directly from flow-level network data, where each flow aggregates packets sharing source, destination, and protocol attributes. Network-level anomaly detection thus provides visibility into a diverse set of attack classes. Reconnaissance attacks such as port and vulnerability scans manifest as repetitive short flows; volumetric floods (UDP, DNS, ICMP) appear as sustained high-throughput flows with abnormal temporal correlations; and stealthy application-layer attacks like Slowloris exhibit long-duration but low-rate deviations. By analyzing traffic at this granularity, AD models can detect both acute, short-term disruptions and gradually evolving threats \cite{DOMENECH2025101631,HAIDER2025111484}.

While anomaly detection has been widely studied in other cyber–physical domains, IoT environments introduce distinctive challenges. IoT traffic is inherently heterogeneous—devices differ in computational capabilities, communication protocols, and sampling rates—yielding highly multivariate traffic, variable sequence lengths, and irregular sampling intervals \cite{Darban2024DeepLearningSurvey}. In addition, IoT deployments are distributed across diverse environments (e.g., smart grids, medical sensors, military automation), which produces severe \emph{domain shift} when models trained on one setting are applied to another \cite{DACAD}. Finally, the scarcity of labeled anomalies in IoT networks makes purely supervised approaches impractical \cite{RAHMAN2025100082,Darban2024DeepLearningSurvey}.

Machine learning–based AD has emerged as a natural solution, modeling complex nonlinear dependencies without heavy feature engineering \cite{HAIDER2025111484}. Within this space, two dominant paradigms exist: reconstruction-based methods (e.g., autoencoders) that flag large reconstruction errors, and prediction-based methods that flag large forecasting residuals; both exploit temporal correlations in multivariate traffic \cite{Darban2024DeepLearningSurvey}. Recent time-series work also leverages contrastive objectives to improve representation quality for downstream detection \cite{ijcai2024p548}.

Among these, sequence models such as LSTM-based autoencoders capture both short- and long-range temporal dependencies, and Variational Autoencoders (VAEs) extend this framework by probabilistically modeling latent spaces to improve generalization in unsupervised settings \cite{GUAN2024103877}. Despite these advantages, commonly used VAE-based approaches in IoT still face two critical limitations: (i) many assume fixed-length input sequences, limiting robustness to real-world traffic variability, and (ii) they often lack mechanisms to transfer knowledge across heterogeneous domains without retraining on target data \cite{Darban2024DeepLearningSurvey,DACAD}.

Effective AD in IoT networks therefore requires models that can (a) accept variable-length flows to capture both rapid and gradual deviations, (b) remain robust under domain shift without fine-tuning on new data, and (c) operate in zero-shot conditions without access to labeled target samples \cite{Darban2024DeepLearningSurvey,DACAD,RAHMAN2025100082}. Few existing approaches simultaneously address these requirements; contrastive learning and domain adaptation are promising ingredients but are rarely combined with strict zero-shot constraints in flow-based IoT settings \cite{ijcai2024p548,DACAD}.

To fill this gap, we propose a zero-shot anomaly detection framework that integrates adversarial domain adaptation and contrastive representation learning into a sequence-driven LSTM–VAE architecture. Our method enforces domain-invariant latent representations while structuring the latent space to capture semantic relationships between flows. Unlike prior work, it supports variable-length sequences and generalizes to unseen domains without requiring raw data, labels, or retraining, making it suitable for dynamic, privacy-sensitive, and continually evolving IoT ecosystems.

\section{Related Work}


\subsection{Supervised and Semi-Supervised AD}
Label-rich models generally achieve strong in-domain accuracy but require continuous annotation and re-labeling as traffic evolves. ST-VAE~\cite{sc_stvae} combines spatiotemporal VAE encoding with supervised contrastive learning and entropy-aware sampling to mitigate concept drift; however, its reliance on labeled feedback limits deployment in label-scarce environments. Transformer-based detectors such as SwinIoT~\cite{swin_iot} tailor hierarchical vision Transformers and SLAA modules to structured visual inputs at the edge, yet the specialization to vision signals constrains applicability to general multivariate telemetry. Classical supervised pipelines (e.g., TF–IDF over system calls~\cite{vsm_syscall}) offer interpretability and low cost but lack temporal modeling and do not transfer across unseen domains. SDN-integrated policies~\cite{sdn_iiot} optimize network-layer behavior with tight control loops, but policy coupling hampers portability to heterogeneous infrastructures.

\subsection{Unsupervised AD}
Unsupervised approaches remove the label dependency but still struggle with domain shift. DUDetector~\cite{dudetector} combines Transformer modules and dual-resolution autoencoding for fine-grained segmentation, at the expense of higher compute budgets. EvoAAE~\cite{evoaae} augments adversarial training with PSO-driven architecture search, but training stability can be fragile. Graph-based models such as DPGLAD~\cite{dpglad} exploit dynamic diffusion over evolving topologies, yet require domain-specific graph construction and extensive feature engineering. Lightweight statistical detectors (e.g., moment-based features~\cite{smdt}) are efficient but typically lack the representational capacity for heterogeneous IoT domains. Platform-bound pipelines like MindFlow~\cite{mindflow} (CNN+BiLSTM on NF-BoT-IoT) highlight temporal modeling benefits but limit portability across toolchains and protocols. Federated variants~\cite{light_ad,sfo_fl_if} improve privacy via decentralized training, though they often depend on synchronized participation and sensitive hyperparameters; practical deployments face stragglers and non-IID data issues.

\subsection{Few-/Zero-Shot AD and Domain Shift Without Retraining}
A growing line of work targets adaptation with few or no labels in the target domain. Typical strategies rely on (i) metric learning (contrastive/triplet) to shape a semantically meaningful latent space, and (ii) domain alignment (adversarial or discrepancy-based) to encourage invariance. However, many methods still \emph{fine-tune} on target data (even unlabeled), which weakens the zero-shot guarantee and adds operational complexity. Our framework is designed for \emph{strict zero-shot} inference on the target domain: the final phase computes no loss on target data and requires no target-side retraining, while contrastive and adversarial objectives (learned on source/auxiliary domains) yield a latent geometry that generalizes across domains. 


\subsection{Positioning and Distinctions}
\textbf{Key differentiators of our approach} relative to prior work:
\begin{itemize}
    \item \textbf{Fully unsupervised source-side training with strict zero-shot target inference:} No labels are required from source or target, and the final phase performs decisions without any target loss or fine-tuning.
    \item \textbf{Dual latent alignment:} A combination of contrastive learning (semantic structuring) and adversarial domain alignment (invariance) produces transferrable representations under domain shift.
    \item \textbf{No handcrafted features or graph construction:} The method is protocol-agnostic and avoids bespoke feature engineering or topology building.
    \item \textbf{Privacy-aware operation:} No raw feature transfer from target sites is required; the model transfers solely via learned parameters trained on source/auxiliary domains.
\end{itemize}

\subsection{Comparison With Representative Methods}
To highlight axes that matter in deployment, Table~\ref{tab:rw_matrix} compares representative works along: supervision level, data modality (multivariate flows), support for variable-length sequences, explicit handling of domain shift, zero-/few-shot readiness, and retraining needs on new domains.%
\footnote{``NR'' denotes \emph{not reported} or unspecified in the cited source.}

\begin{table*}[t]
\centering
\caption{Comparison of representative IoT anomaly detection methods.}
\label{tab:rw_matrix}
\scriptsize
\resizebox{\textwidth}{!}{%
\begin{tabular}{lccccccp{4.5cm}}
\toprule
\textbf{Method} & \textbf{Supervision} & \textbf{Multivar.} & \textbf{Var. Len.} & \textbf{Domain Shift} & \textbf{Zero/Few} & \textbf{Retrain?} & \textbf{Notes} \\
\midrule
ST-VAE~\cite{sc_stvae} & Sup. & Yes & NR & Partial & Few-shot & Often & Supervised contrastive + spatiotemporal VAE. \\
SwinIoT~\cite{swin_iot} & Sup. & Vision & NR & No & No & Yes & Hierarchical Transformers for edge vision only. \\
TF–IDF Syscalls~\cite{vsm_syscall} & Sup. & Syscall text & No & No & No & Yes & Lightweight, but weak temporal modeling. \\
SDN-IIoT Policy~\cite{sdn_iiot} & Sup./Rules & Net. flows & NR & No & No & Yes & Tight SDN coupling; limited portability. \\
DUDetector~\cite{dudetector} & Unsup. & Yes & NR & Partial & No & Possibly & Transformer + dual-res AE; high compute. \\
EvoAAE~\cite{evoaae} & Unsup. (adv.) & Yes & NR & Partial & No & Possibly & Adversarial AE + PSO NAS; unstable. \\
DPGLAD~\cite{dpglad} & Unsup. (graph) & Graph & NR & Partial & No & Possibly & Needs domain-specific graph construction. \\
Stat. Moments~\cite{smdt} & Sup./Heur. & Yes (hand) & NR & No & No & Yes & Efficient, limited to handcrafted features. \\
MindFlow~\cite{mindflow} & Sup. & Yes & NR & No & No & Yes & CNN+BiLSTM; platform-dependent. \\
Fed. GAN/Ensemble~\cite{light_ad} & Fed. & Yes & NR & Partial & Few-shot & Sometimes & FL; sync/global aggregation required. \\
Fed. IF Opt.~\cite{sfo_fl_if} & Fed. & Yes & NR & Partial & Few-shot & Sometimes & FL, hyperparam-sensitive. \\
DACAD~\cite{DACAD} & Unsup./DA & Yes & NR & Yes & Zero/Few & Often & Contrastive domain adaptation for multivariate time-series AD; strong baseline but not designed around destination-centric IoT flow segmentation, variable-length LSTM-VAE reconstruction, or strict no-target-loss inference. \\
\textbf{This work} & \textbf{Unsup.} & \textbf{Yes} & \textbf{Yes} & \textbf{Yes} & \textbf{Zero/Few} & \textbf{No} & Strict zero-shot inference; dual alignment; validated on 6 datasets. \\
\bottomrule
\end{tabular}%
}
\end{table*}

\section{System Model}

\begin{figure*}[t]
    \centering
    \includegraphics[width=0.65\textwidth]{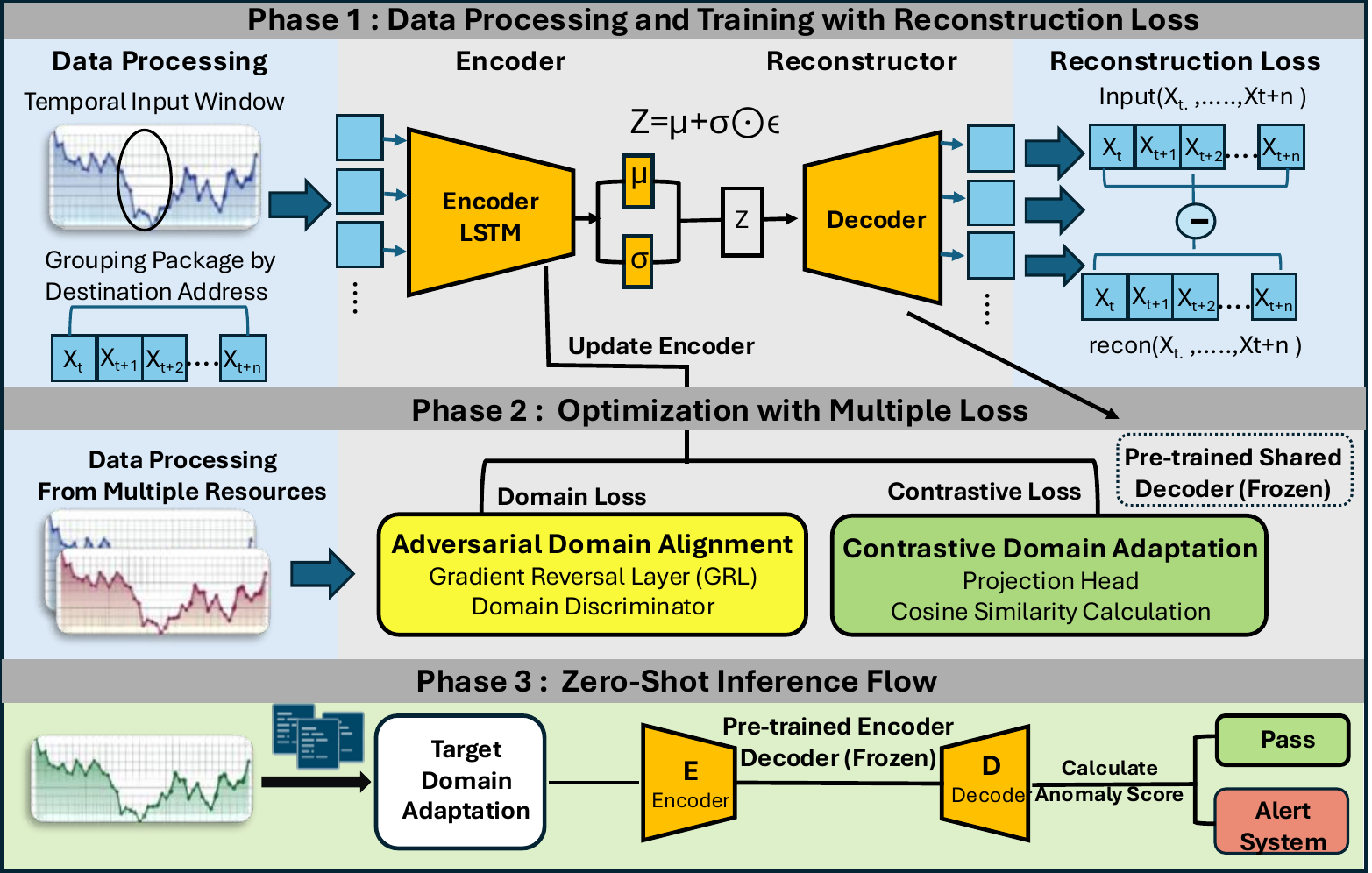}
    \caption{Overall architecture of the proposed Contrastive Adversarially-Adaptive LSTM-VAE framework. The model is trained using two unlabeled source datasets to learn domain-invariant and semantically structured latent representations. Adversarial learning minimizes the domain discrepancy through a gradient reversal layer and domain classifier, while contrastive loss refines the latent space geometry. Adaptors normalize cross-domain input/output formats. At inference, the model generalizes to the unseen target domain in a zero-shot manner using reconstruction error and latent deviations for anomaly detection.}
    \label{fig:model_architecture}
\end{figure*}

In this work, we propose the \textit{Contrastive Adversarially-Adaptive LSTM-VAE}, a novel framework tailored for multivariate anomaly detection in IoT and networked environments. The overall architecture of the proposed model is illustrated in Figure~\ref{fig:model_architecture}.

Our framework builds upon a Variational Autoencoder (VAE) augmented with Long Short-Term Memory (LSTM) layers to effectively capture temporal dependencies in multivariate time-series data. To address domain shifts across different network traffic distributions, we introduce lightweight input and output adaptor layers that serve as transformation modules, aligning feature representations between source and target domains.

As shown in Figure~\ref{fig:model_architecture}, the model is trained on two distinct, unlabeled network traffic datasets. The goal is to learn representations that are invariant to dataset-specific characteristics while capturing fundamental patterns in temporal dynamics and cross-feature correlations. This enables the model to generalize based on intrinsic structure rather than domain-dependent artifacts.

The training procedure begins with unsupervised pretraining on a large-scale source dataset, where the Variational Autoencoder (VAE) learns to reconstruct benign traffic patterns and encode variable-length input sequences into compact latent representations. To minimize the domain gap between the source and auxiliary datasets, adversarial domain alignment is applied. Specifically, a domain discriminator is trained to identify the origin of encoded samples, while the encoder is concurrently optimized in an adversarial manner to mislead the discriminator, thereby promoting the learning of domain-invariant latent features.

In addition, to develop a more generalizable model that is also capable of zero-shot adaptation, it is essential to enhance the semantic structure of the latent space. This facilitates the model’s ability to extract and retain meaningful knowledge even when input sequences vary in length and no labeled or unlabeled samples from the target domain are available during training. To achieve this, we incorporate a contrastive learning objective, which encourages the model to learn discriminative and semantically meaningful representations under such constraints.
Specifically, the encoder is guided to cluster similar patterns (positive pairs) and separate dissimilar ones (negative pairs) based on cosine similarity, even when they originate from different domains. This dual training strategy—adversarial for alignment and contrastive for structure—produces a latent space that is both generalizable and discriminative.

After training, the model is deployed to a previously unseen target domain in a zero-shot manner, meaning no retraining or target-domain labels are used. Anomalies are detected based on reconstruction errors and deviations from the latent distribution learned during training on source domains.

To provide a structured overview of our approach, we first formalize the anomaly detection problem and the challenges of domain adaptation in time-series data. We then detail the architecture and components of our framework, including the LSTM-VAE core, domain-specific adaptor layers, and the integration of adversarial and contrastive learning mechanisms.

\subsection{Problem Definition}
\label{subsec:problem_definition}

We model IoT network traffic as a multivariate time series, where each observation is a feature vector recorded at a discrete time step. Formally, a traffic sequence is denoted as
\[
\mathbf{X} = (x_1,x_2,\ldots,x_T) \in \mathbb{R}^{T \times d},
\]
where $T$ is the sequence length and $d$ is the number of traffic features. In flow-level network monitoring, these features may include packet counts, flow duration, inter-arrival statistics, throughput, byte counts, protocol indicators, or other traffic descriptors. A collection of such features over time forms a multivariate sequence that captures both temporal dependencies and cross-feature correlations in network behavior.

This work addresses anomaly detection under heterogeneous cross-domain IoT deployments. Let $\mathcal{D}_{s}$ denote a primary source domain, $\mathcal{D}_{a}$ denote an auxiliary source domain used for domain alignment, and $\mathcal{D}_{t}$ denote a held-out target domain. The source and auxiliary domains may come from different network environments, devices, protocols, or attack distributions. The target domain represents an unseen deployment whose data distribution is not available during training. The goal is to learn an anomaly detector from $\mathcal{D}_{s}$ and $\mathcal{D}_{a}$ that generalizes to $\mathcal{D}_{t}$ without target-domain labels, target-domain fine-tuning, or target-domain loss computation.

The proposed framework follows a strict zero-shot protocol. In the first stage, an LSTM-based VAE is trained on the primary source domain to model normal temporal behavior through sequence reconstruction and latent-space regularization. In the second stage, samples from the source and auxiliary domains are used to improve transferability through adversarial domain alignment and contrastive representation learning. The adversarial objective encourages the encoder to suppress domain-specific artifacts between $\mathcal{D}_{s}$ and $\mathcal{D}_{a}$, while the contrastive objective structures the latent space by pulling similar temporal patterns closer and pushing dissimilar patterns apart. Importantly, the held-out target domain $\mathcal{D}_{t}$ is not used in either stage.

After training, the learned encoder--decoder model is deployed directly on $\mathcal{D}_{t}$. For each target sequence, the model computes an anomaly score from reconstruction error and latent deviation relative to the source-learned normal-behavior representation. A sequence is classified as anomalous when its score exceeds a predefined threshold. Target-domain labels are used only for final evaluation metrics and are never used for training, alignment, threshold selection, or model adaptation. This formulation enables privacy-preserving and deployment-oriented anomaly detection in heterogeneous IoT environments where collecting labeled target data or retraining the model for every new domain is impractical.

\subsection{Sequence-Based VAE}

We employ a sequence-based Variational Autoencoder (VAE) architecture to model multivariate time-series data in IoT environments. The goal is to encode an observed sequence $x$ into a latent representation $z$ while retaining its essential structural and temporal features. The VAE models the joint probability distribution of observed variables $x$ and latent variables $z$, assuming a standard Gaussian prior over $z$:
\begin{equation}
p(z) = \mathcal{N}(z; 0, I)
\end{equation}
The data $x$ is generated conditionally through a neural decoder $p_\theta(x|z)$, but the true posterior $p(z|x)$ is intractable. To approximate it, we use an encoder $q_\phi(z|x)$ parameterized as a Gaussian with learned mean and variance:
\begin{equation}
q_\phi(z|x) = \mathcal{N}(z; \mu(x), \sigma^2(x))
\end{equation}
\noindent where the approximate posterior takes the form:
\begin{equation}
q_\phi(z|x) = \frac{1}{\sqrt{2\pi \sigma^2(x)}} \exp\left( - \frac{(z - \mu(x))^2}{2\sigma^2(x)} \right)
\end{equation}
To enable backpropagation through sampling, we apply the reparameterization trick:
\begin{equation}
z = \mu(x) + \sigma(x) \cdot \epsilon, \quad \epsilon \sim \mathcal{N}(0, I)
\end{equation}
\noindent In our approach, each input sequence $x = \{x_1, x_2, \ldots, x_T\}$, representing a series of network traffic observations, is fed into the model. Instead of treating each time step independently, we use an LSTM to capture temporal dependencies across the sequence.
The encoder LSTM processes the entire sequence and computes the final hidden state $h_T$, which summarizes it. The latent variables are derived as:
\begin{align}
h_t &= f_{\text{LSTM}}(h_{t-1}, x_t) \\
\mu_\phi(x) &= W_\mu h_T + b_\mu \\
\log \sigma^2_\phi(x) &= W_\sigma h_T + b_\sigma
\end{align}
\noindent The latent vector $z$ is sampled using the learned parameters:
\begin{equation}
z = \mu_\phi(x) + \sigma_\phi(x) \cdot \epsilon, \quad \epsilon \sim \mathcal{N}(0, I)
\end{equation}
The decoder is also an LSTM network and reconstructs the sequence one step at a time. Given $z$ and the previously reconstructed step $\hat{x}_{t-1}$, each next step is generated as:
\begin{equation}
\hat{x}_t = f_{\text{decoder}}(z, \hat{x}_{t-1})
\end{equation}
This sequence-based formulation allows the model to learn both short- and long-term temporal patterns in IoT traffic data. By encoding the full sequence into a latent space and reconstructing it iteratively, the model builds a nuanced understanding of normal traffic behavior. Deviations in the reconstruction signal potential anomalies, enabling effective detection of both sudden and gradual attacks across varied environments.

\subsection{Variable-Length Sequence Adaptation}
\label{subsec:variable_length}

Real IoT traffic rarely arrives as uniform fixed-length sequences. Flow durations vary across devices, protocols, sampling rates, and attack behaviors. A fixed-window detector can therefore introduce two practical problems: short flows may be padded with little useful context, while long flows may be truncated before slow or gradual anomalies become visible. Both effects can degrade detection performance and delay operational response.

To address this issue, the proposed framework supports variable-length traffic sequences. Each sequence is encoded by the LSTM encoder into a compact latent representation, allowing the model to summarize temporal behavior without requiring all inputs to share the same duration. This makes the detector more suitable for heterogeneous IoT environments, where attacks may appear as short bursts, repeated scans, or long-duration low-rate deviations.

Because shorter sequences may contain limited temporal context, variable-length modeling is combined with adversarial and contrastive latent learning. Adversarial alignment reduces domain-specific artifacts between the primary and auxiliary source domains, while contrastive learning encourages similar temporal patterns to remain close in the latent space. Together, these mechanisms help the model preserve useful behavioral structure across different sequence lengths and deployment domains.

\subsection{Adversarial and Contrastive Learning for Domain-Invariant Representations}
\label{subsec:adv_contrastive}

To improve cross-domain generalization, we integrate adversarial domain alignment and cosine-based contrastive learning into the latent-space training stage. Let $\mathcal{D}_{s}$ denote the primary source domain and $\mathcal{D}_{a}$ denote the auxiliary source domain used for alignment. The held-out target domain $\mathcal{D}_{t}$ is not used during this stage. The goal is to learn an encoder $E$ that suppresses domain-specific artifacts between $\mathcal{D}_{s}$ and $\mathcal{D}_{a}$ while preserving temporal structures that are useful for anomaly detection in unseen deployments.

\subsubsection{Adversarial Domain Alignment}

We introduce a domain discriminator $D$ that predicts whether a latent representation originates from the primary source domain $\mathcal{D}_{s}$ or the auxiliary source domain $\mathcal{D}_{a}$. The discriminator is trained to distinguish the two domains, while the encoder is trained through a Gradient Reversal Layer (GRL) to make the two latent distributions difficult to separate. This adversarial objective encourages the encoder to remove domain-specific nuisance factors and retain features that are more transferable across environments.

The adversarial loss is defined as
\begin{equation}
\begin{aligned}
\mathcal{L}_{\mathrm{adv}}(E,D) =
&-\mathbb{E}_{x_s \sim \mathcal{D}_{s}}
\left[\log D(E(x_s))\right]  \\
&-\mathbb{E}_{x_a \sim \mathcal{D}_{a}}
\left[\log \left(1-D(E(x_a))\right)\right],
\end{aligned}
\end{equation}
where $x_s$ and $x_a$ denote samples from the primary and auxiliary source domains, respectively. The discriminator minimizes $\mathcal{L}_{\mathrm{adv}}$, whereas the encoder receives the reversed gradient through the GRL, which makes the latent representations less domain-identifiable.

\subsubsection{Cosine-Based Contrastive Learning}

Adversarial alignment alone may remove domain-specific information but does not necessarily impose a useful semantic geometry on the latent space. Therefore, we also apply cosine-based contrastive learning to encourage similar temporal patterns to remain close while dissimilar patterns are separated. For each anchor representation $z_i^a$, we define a positive representation $z_i^p$ and a negative representation $z_i^n$. The contrastive loss is

\begin{equation}
\mathcal{L}_{\mathrm{con}} =
\frac{1}{N}\sum_{i=1}^{N}
\left[
1-\cos(z_i^a,z_i^p)
+
\max\left(0,\cos(z_i^a,z_i^n)-m\right)
\right],
\end{equation}
where $\cos(\cdot,\cdot)$ denotes cosine similarity and $m$ is a margin controlling the minimum desired separation between anchor-negative pairs. This objective improves latent-space organization by reducing intra-pattern variation and increasing separation between dissimilar temporal behaviors.

\subsubsection{Combined Objective}

The final phase--2 training objective combines adversarial alignment and contrastive structuring:
\begin{equation}
\mathcal{L}_{\mathrm{phase2}}
=
\lambda_{\mathrm{adv}}\mathcal{L}_{\mathrm{adv}}
+
\lambda_{\mathrm{con}}\mathcal{L}_{\mathrm{con}},
\end{equation}
where $\lambda_{\mathrm{adv}}$ and $\lambda_{\mathrm{con}}$ control the relative strength of domain confusion and contrastive latent organization. The target domain $\mathcal{D}_{t}$ is excluded from this optimization and is used only during final zero-shot evaluation. Thus, the learned representation is encouraged to be transferable without requiring target-domain labels, target-domain loss computation, or target-side fine-tuning.

\subsection{Domain-Specific Adaptors for Variability Handling}
\label{subsec:adaptors}

IoT datasets often differ in feature scale, feature ordering, measurement units, protocol composition, and device behavior. Directly feeding such heterogeneous inputs into a shared sequence model can cause the encoder to learn dataset-specific artifacts rather than transferable temporal structure. To reduce this effect, we introduce lightweight domain-specific adaptor layers around the shared LSTM-VAE.

For a sample $x_i^{(k)}$ from domain $\mathcal{D}_k$, where $k\in\{s,a\}$ during training, the encoder adaptor $A_e^{(k)}$ maps the input sequence into a shared feature space:
\begin{equation}
\tilde{x}_i^{(k)} = A_e^{(k)}(x_i^{(k)}).
\end{equation}
The transformed sequence $\tilde{x}_i^{(k)}$ is then passed to the shared LSTM encoder, which produces the latent representation $z_i^{(k)}$. This design allows the source and auxiliary domains to be normalized before entering the shared latent model.

Similarly, after decoding, a lightweight decoder adaptor $A_d^{(k)}$ maps the decoder output back to the corresponding domain-specific feature space:
\begin{equation}
\hat{x}_i^{(k)} = A_d^{(k)}(\hat{\tilde{x}}_i^{(k)}).
\end{equation}
The adaptor layers therefore act as interface modules that absorb domain-specific feature variability, while the shared LSTM-VAE focuses on learning transferable temporal dynamics.

During strict zero-shot inference, no target-specific adaptor is trained using target data. The learned shared encoder--decoder is applied directly to the held-out target domain, and the anomaly decision is computed from the reconstruction and latent-deviation score. This preserves the zero-shot protocol while still allowing the model to benefit from source--auxiliary adaptor-based alignment during training.

\subsection{Loss Functions and Training Phases}
\label{subsec:loss_training}

The proposed framework is trained in two stages. Phase~1 learns a reconstruction-based normal-behavior model from the primary source domain. Phase~2 improves cross-domain transferability by aligning the primary source domain with an auxiliary source domain using adversarial and contrastive objectives. The held-out target domain is excluded from both training stages and is used only for final zero-shot evaluation.

\subsubsection{Phase 1: VAE Pretraining on the Primary Source Domain}

In the first stage, the LSTM-VAE is pretrained on the primary source domain $\mathcal{D}_{s}$ to model normal temporal behavior. Given an input sequence $x_i$, the encoder produces the approximate posterior parameters $\mu_i$ and $\sigma_i$, samples a latent vector $z_i$ through the reparameterization trick, and the decoder reconstructs the sequence as $\hat{x}_i$. The reconstruction loss is defined as
\begin{equation}
\mathcal{L}_{\mathrm{rec}}
=
\frac{1}{N}\sum_{i=1}^{N}
\left\|x_i-\hat{x}_i\right\|_2^2 .
\end{equation}

The Kullback--Leibler divergence regularizes the approximate posterior $q_{\phi}(z|x)$ toward the standard Gaussian prior $p(z)=\mathcal{N}(0,I)$:
\begin{equation}
\mathcal{L}_{\mathrm{KL}}
=
-\frac{1}{2}\sum_{j=1}^{d_z}
\left(
1+\log \sigma_{ij}^{2}
-\mu_{ij}^{2}
-\sigma_{ij}^{2}
\right),
\end{equation}
where $d_z$ is the latent dimension. The Phase~1 objective is
\begin{equation}
\mathcal{L}_{\mathrm{phase1}}
=
\lambda_{\mathrm{rec}}\mathcal{L}_{\mathrm{rec}}
+
\lambda_{\mathrm{KL}}\mathcal{L}_{\mathrm{KL}},
\end{equation}
where $\lambda_{\mathrm{rec}}$ and $\lambda_{\mathrm{KL}}$ control the relative contributions of reconstruction fidelity and latent regularization.

\subsubsection{Phase 2: Source--Auxiliary Alignment}

In the second stage, we improve transferability using the primary source domain $\mathcal{D}_{s}$ and an auxiliary source domain $\mathcal{D}_{a}$. The target domain $\mathcal{D}_{t}$ is not used in this phase. A domain discriminator $D$ attempts to distinguish whether an encoded representation comes from $\mathcal{D}_{s}$ or $\mathcal{D}_{a}$, while the encoder $E$ is optimized through a Gradient Reversal Layer (GRL) to make the two domains difficult to separate. The adversarial domain loss is
\begin{equation}
\begin{aligned}
\mathcal{L}_{\mathrm{GRL}}
=
&-\mathbb{E}_{x_s\sim \mathcal{D}_{s}}
\left[\log D(E(x_s))\right] \\
&-\mathbb{E}_{x_a\sim \mathcal{D}_{a}}
\left[\log\left(1-D(E(x_a))\right)\right],
\end{aligned}
\end{equation}
where $x_s$ and $x_a$ are samples from the primary and auxiliary source domains, respectively.

To further organize the latent space, we apply cosine-based contrastive learning. For each anchor representation $z_i^a$, let $z_i^p$ denote a positive representation and $z_i^n$ denote a negative representation. The contrastive loss is
\begin{equation}
\mathcal{L}_{\mathrm{con}}
=
\frac{1}{N}\sum_{i=1}^{N}
\left[
1-\cos(z_i^a,z_i^p)
+
\max\left(0,\cos(z_i^a,z_i^n)-m\right)
\right],
\end{equation}
where $\cos(\cdot,\cdot)$ denotes cosine similarity and $m$ is the margin that controls anchor-negative separation.

The Phase~2 objective is then
\begin{equation}
\mathcal{L}_{\mathrm{phase2}}
=
\lambda_{\mathrm{GRL}}\mathcal{L}_{\mathrm{GRL}}
+
\lambda_{\mathrm{con}}\mathcal{L}_{\mathrm{con}},
\end{equation}
where $\lambda_{\mathrm{GRL}}$ and $\lambda_{\mathrm{con}}$ balance adversarial domain confusion and contrastive latent structuring.

\subsubsection{Phase 3: Strict Zero-Shot Inference}

After training, the encoder and decoder are frozen and deployed directly on the held-out target domain $\mathcal{D}_{t}$. No target-domain labels, target-domain loss, target-domain fine-tuning, or target-domain model selection are used. For a target sequence $x_t$, the model computes an anomaly score from the reconstruction error and latent deviation:
\begin{equation}
s(x_t)
=
\alpha \left\|x_t-\hat{x}_t\right\|_2^2
+
(1-\alpha)\left\|z_t-\bar{z}_{s}\right\|_2^2,
\end{equation}
where $z_t=E(x_t)$, $\bar{z}_{s}$ is the mean latent representation estimated from normal source-domain training samples, and $\alpha\in[0,1]$ controls the relative contribution of reconstruction and latent deviation. A target sequence is classified as anomalous when $s(x_t)$ exceeds a predefined threshold. Target labels are used only to compute final evaluation metrics.

\subsection{Evaluation Datasets} 

In this study, we utilize six distinct IoT-related datasets originating from diverse application domains—namely, industrial IoT, general-purpose enterprise networks, military automation, academic testbeds, and smart home automation—to comprehensively evaluate the effectiveness and generalizability of our proposed anomaly detection framework.

\begin{itemize}
    \item \textbf{CICIDS 2018 Dataset:} The CICIDS 2018 dataset \cite{CSE_CIC_IDS2018}, curated by the Canadian Institute for Cybersecurity (CIC) in collaboration with the Communications Security Establishment (CSE), was generated using a simulated enterprise network setup. It spans ten days of realistic network traffic, totaling over 16 million flow-based entries. Approximately 83\% of the data is benign, while 17\% comprises various attack categories including brute force, denial-of-service (DoS), botnet activity, web attacks, and infiltration. Each record contains more than 80 traffic flow features extracted via CICFlowMeter. This dataset serves as a comprehensive benchmark for evaluating intrusion detection systems in general network environments. All recorded features are monitored to examine the robustness of our method under diverse traffic behaviors, enabling an in-depth assessment of model adaptability across heterogeneous network conditions.
    
    \item \textbf{WUSTL-IIOT-2021 Dataset:} The WUSTL-IIOT-2021 dataset \cite{WUSTL_IIoT_2021} originates from an Industrial IoT (IIoT) testbed designed to monitor water storage tanks. Spanning 53 hours of operation, it includes 1,194,464 entries, capturing critical network metrics such as packet drops and flow durations. The dataset is predominantly benign (93\%), with 7\% of records corresponding to attack types such as denial-of-service (89.98\%), reconnaissance (9.46\%), SQL injection (0.31\%), and backdoor access (0.25\%). Consistent with prior studies, we track all extracted features to analyze the effectiveness of the proposed method when applied to industrial traffic with varying characteristics.
    
    \item \textbf{ACI-IoT-2023 Dataset:} The ACI-IoT-2023 dataset \cite{Bastian2023_ACIIoT} was collected within a simulated military IoT environment that replicates real-world home automation scenarios. It contains 742,758 traffic entries recorded over five days, of which 95\% are malicious. The attack composition includes reconnaissance (74\%), brute force (1\%), and denial-of-service (25\%), encompassing subcategories such as scanning and sweeping. This dataset offers valuable insights into the cybersecurity vulnerabilities inherent in military-oriented IoT systems. We utilize all available features to evaluate the proposed method’s performance in handling traffic patterns with distinct operational contexts.

    \item \textbf{CICIDS 2017 Dataset:} A predecessor to CICIDS 2018, this dataset \cite{Sharafaldin2018_CICIDS2017} includes both benign and common, up-to-date attacks. It contains 2,830,743 total flow entries, of which approximately 80.3\% are benign. The remaining 19.7\% of malicious traffic is composed of several attack types. The most prevalent attacks are Denial-of-Service (DoS/DDoS), accounting for the vast majority of malicious instances, followed by Port Scanning. Other implemented attacks include Brute Force (FTP and SSH), Web Attacks (Brute Force, XSS, SQL Injection), Infiltration, Botnet, and Heartbleed. The dataset is used in source–target–adversary combinations to evaluate robustness across different temporal captures and protocol variations within CIC environments.

    \item \textbf{UNSW-NB15 Dataset:} Developed by the Australian Cyber Security Centre, this dataset \cite{Moustafa2015_UNSWNB15} includes both real benign and synthetic attack flows from a total of 2,540,044 records. Approximately 87.4\% of the traffic is benign, while the other 12.6\% is malicious. The attack composition consists of nine types, dominated by Generic attacks (67.1\%) and Exploits (13.9\%). The remaining categories are Fuzzers (7.5\%), DoS (5.1\%), Reconnaissance (4.3\%), Analysis (0.8\%), Backdoors (0.7\%), Shellcode (0.5\%), and Worms (0.1\%). It is employed as both a source and target to evaluate cross-protocol and attack-style generalization.

    \item \textbf{TON-IoT Dataset:} Provided by the Cyber Range Lab at UNSW, TON-IoT \cite{Moustafa2020_ToNIoT} is a smart home-oriented dataset with over 22 million telemetry and network event records. Within the network traffic portion, approximately 64.8\% of entries are benign, and 35.2\% are malicious. The malicious traffic is heavily skewed towards reconnaissance activities, with Scanning attacks making up 48.7\% of all malicious events. Denial-of-Service is also highly prevalent, with DDoS (29.0\%) and DoS (21.6\%) attacks forming the next largest categories. The dataset also contains smaller numbers of Backdoor (0.34\%), Injection (0.19\%), Ransomware (0.10\%), XSS (0.04\%), Password (0.01\%), and Man-in-the-Middle (MITM) attacks. It is used as a target domain in several experiments, revealing generalization challenges in more constrained settings.
\end{itemize}

Together, these six datasets provide a heterogeneous evaluation suite spanning enterprise network traffic, industrial IoT, military-oriented IoT, academic intrusion-detection testbeds, and smart-home IoT environments. CICIDS2017 and CICIDS2018 represent general-purpose enterprise network settings with diverse benign and malicious flow patterns; WUSTL-IIoT-2021 captures industrial IoT traffic with operationally structured benign behavior and a smaller set of attack classes; ACI-IoT-2023 reflects a military-oriented IoT environment with a highly attack-dominant distribution; UNSW-NB15 introduces additional protocol and attack-style diversity; and TON-IoT provides smart-home IoT traffic with substantial reconnaissance and denial-of-service activity.

In our evaluation framework, these datasets are used in source--auxiliary--target combinations to test cross-domain generalization under strict zero-shot conditions. For each experiment, the model is trained using a primary source domain $\mathcal{D}_{s}$ and an auxiliary source domain $\mathcal{D}_{a}$ for adversarial and contrastive alignment, while the target domain $\mathcal{D}_{t}$ is held out entirely during training. No target-domain samples, labels, losses, or fine-tuning steps are used before inference. The target dataset is used only for final metric computation. This protocol allows us to evaluate both successful transfer cases and failure modes caused by domain shift, class imbalance, and insufficient source-domain coverage.

\section{Model Evaluation and Analysis}


\subsection{Experimental Protocol}

In all experiments, the held-out target domain is used only for final evaluation. Target labels are not used during model training, adversarial alignment, contrastive learning, threshold selection, or model selection. Performance is reported using accuracy, Matthews correlation coefficient (MCC), recall, precision, F1-score, and false-positive rate (FPR). This protocol is intended to measure strict zero-shot transfer rather than target-adaptive performance.



\begin{figure}[h!]
\centering
\includegraphics[width=0.4\textwidth]{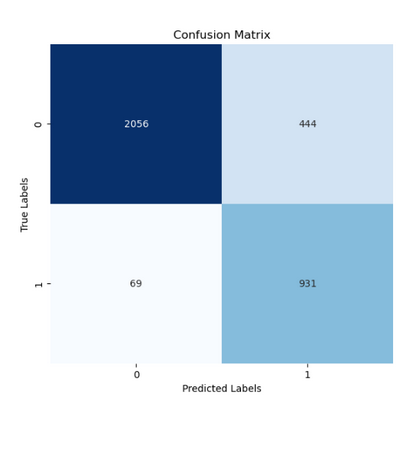}
\caption{Confusion matrix for transfer scenario: CICIDS 2018 (source) $\rightarrow$ ACI-IoT-2023 (target)}
\label{fig:conf_matrix_cicids}
\end{figure}

\begin{figure}[h!]
\centering
\includegraphics[width=0.4\textwidth]{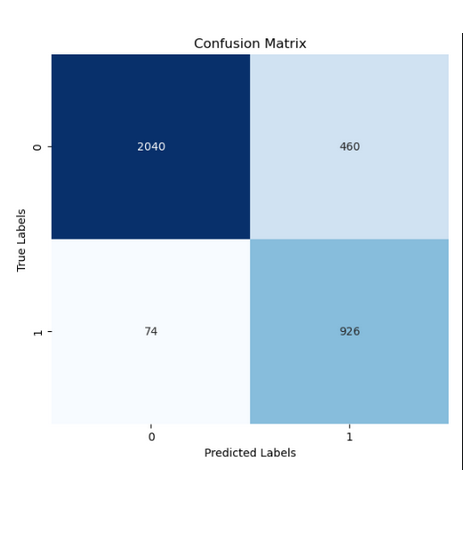}
\caption{Confusion matrix for transfer scenario: WUSTL-IIoT-2021 (source) $\rightarrow$ ACI-IoT-2023 (target)}
\label{fig:conf_matrix_washington}
\end{figure}

\begin{figure}[h!]
\centering
\includegraphics[width=0.4\textwidth]{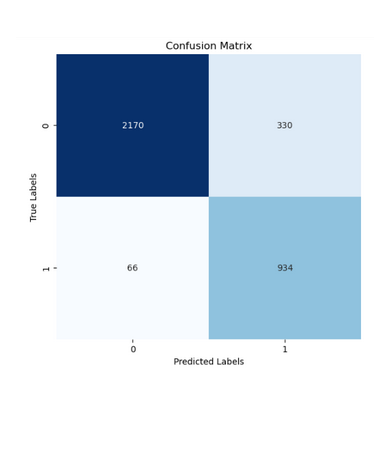}
\caption{Confusion matrix for transfer scenario: CICIDS 2018 + WUSTL-IIoT-2021 (source) $\rightarrow$ ACI-IoT-2023 (target)}
\label{fig:conf_matrix_army}
\end{figure}

The confusion matrices in Figs.~\ref{fig:conf_matrix_cicids}--\ref{fig:conf_matrix_army} reinforce these observations. In the CICIDS-only scenario, true positives are generally high, indicating good recognition of abnormal patterns. However, moderate false positives suggest some overgeneralization to benign classes. The WUSTL-only model achieves a similar true negative rate but suffers from increased false positives and false negatives, due to the absence of explicit attack semantics during training. The combined scenario demonstrates the most balanced classification performance with minimal false negatives (66) and false positives (330), aligning with its top accuracy. This confirms that the adversarial framework effectively aligns heterogeneous source domains to construct a more resilient decision boundary in the target space.


\subsection{Zero-Shot Inference Protocol (No Target Loss)}
In the final phase, the model operates under a strict zero-shot setting on the target domain: no target data (labeled or unlabeled) is used and \emph{no loss function is computed} in this phase. Decisions are made using reconstruction errors and latent deviations only.

\begin{table*}[t]
\centering
\scriptsize
\setlength{\tabcolsep}{6pt} 
\renewcommand{\arraystretch}{1.1} 
\begin{tabular}{llcccccc}
\toprule
\textbf{Source} & \textbf{Target} & \textbf{ACC} & \textbf{MCC} & \textbf{Recall} & \textbf{Precision} & \textbf{F1} & \textbf{FPR} \\
\midrule
ACI+TON                & CICIDS2017  & 1.000 & 1.000 & 1.000 & 1.000 & 1.000 & 0.000 \\
ACI+TON                & CICIDS2018  & 0.833 & 0.707 & 1.000 & 0.750 & 0.857 & 0.333 \\
ACI+TON                & NB15        & 0.812 & 0.584 & 0.781 & 1.000 & 0.877 & 0.000 \\
ACI+WUSTL-IIoT-2021         & CICIDS2017  & 1.000 & 1.000 & 1.000 & 1.000 & 1.000 & 0.000 \\
ACI+WUSTL-IIoT-2021         & CICIDS2018  & 0.854 & 0.737 & 1.000 & 0.781 & 0.877 & 0.304 \\
ACI+WUSTL-IIoT-2021         & NB15        & 0.833 & 0.632 & 0.800 & 1.000 & 0.889 & 0.000 \\
ACI+WUSTL-IIoT-2021         & TON         & 0.479 & 0.227 & 0.889 & 0.250 & 0.390 & 0.615 \\
CICIDS2017+ACI         & CICIDS2018  & 0.875 & 0.769 & 1.000 & 0.812 & 0.897 & 0.273 \\
CICIDS2017+ACI         & NB15        & 0.812 & 0.584 & 0.781 & 1.000 & 0.877 & 0.000 \\
CICIDS2017+ACI         & TON         & 0.479 & 0.227 & 0.889 & 0.250 & 0.390 & 0.615 \\
CICIDS2017+TON         & ACI         & 0.333 & 0.000 & 0.000 & 0.000 & 0.000 & 0.667 \\
CICIDS2017+TON         & CICIDS2018  & 0.958 & 0.913 & 1.000 & 0.938 & 0.968 & 0.111 \\
CICIDS2017+TON         & NB15        & 0.854 & 0.679 & 0.821 & 1.000 & 0.901 & 0.000 \\
CICIDS2017+WUSTL-IIoT-2021  & ACI         & 0.333 & 0.000 & 0.000 & 0.000 & 0.000 & 0.667 \\
CICIDS2017+WUSTL-IIoT-2021  & CICIDS2018  & 0.958 & 0.913 & 1.000 & 0.938 & 0.968 & 0.111 \\
CICIDS2017+WUSTL-IIoT-2021  & NB15        & 0.833 & 0.632 & 0.800 & 1.000 & 0.889 & 0.000 \\
CICIDS2017+WUSTL-IIoT-2021  & TON         & 0.479 & 0.227 & 0.889 & 0.250 & 0.390 & 0.615 \\
CICIDS2018+ACI         & CICIDS2017  & 1.000 & 1.000 & 1.000 & 1.000 & 1.000 & 0.000 \\
CICIDS2018+ACI         & NB15        & 0.833 & 0.632 & 0.800 & 1.000 & 0.889 & 0.000 \\
CICIDS2018+ACI         & TON         & 0.479 & 0.227 & 0.889 & 0.250 & 0.390 & 0.615 \\
CICIDS2018+TON         & ACI         & 0.333 & 0.000 & 0.000 & 0.000 & 0.000 & 0.667 \\
CICIDS2018+TON         & CICIDS2017  & 1.000 & 1.000 & 1.000 & 1.000 & 1.000 & 0.000 \\
\bottomrule
\end{tabular}
\caption{Cross-domain zero-shot results for part 1 of the 44-experiment evaluation. \textbf{Source} lists the two training domains (A+B); \textbf{Target} is the held-out zero-shot domain.}
\label{tab:cross_domain_previous_44_part1}
\end{table*}

\begin{table*}[t]
\centering
\scriptsize
\setlength{\tabcolsep}{6pt}
\renewcommand{\arraystretch}{1.1}
\begin{tabular}{llcccccc}
\toprule
\textbf{Source} & \textbf{Target} & \textbf{ACC} & \textbf{MCC} & \textbf{Recall} & \textbf{Precision} & \textbf{F1} & \textbf{FPR} \\
\midrule
CICIDS2018+TON         & NB15        & 0.812 & 0.584 & 0.781 & 1.000 & 0.877 & 0.000 \\
CICIDS2018+WUSTL-IIoT-2021  & ACI         & 0.333 & 0.000 & 0.000 & 0.000 & 0.000 & 0.667 \\
CICIDS2018+WUSTL-IIoT-2021  & CICIDS2017  & 1.000 & 1.000 & 1.000 & 1.000 & 1.000 & 0.000 \\
CICIDS2018+WUSTL-IIoT-2021  & NB15        & 0.854 & 0.679 & 0.821 & 1.000 & 0.901 & 0.000 \\
CICIDS2018+WUSTL-IIoT-2021  & TON         & 0.479 & 0.227 & 0.889 & 0.250 & 0.390 & 0.615 \\
NB15+ACI               & CICIDS2017  & 1.000 & 1.000 & 1.000 & 1.000 & 1.000 & 0.000 \\
NB15+ACI               & CICIDS2018  & 0.938 & 0.874 & 1.000 & 0.906 & 0.951 & 0.158 \\
NB15+ACI               & TON         & 0.479 & 0.227 & 0.889 & 0.250 & 0.390 & 0.615 \\
NB15+TON               & ACI         & 0.333 & 0.000 & 0.000 & 0.000 & 0.000 & 0.667 \\
NB15+TON               & CICIDS2017  & 1.000 & 1.000 & 1.000 & 1.000 & 1.000 & 0.000 \\
NB15+TON               & CICIDS2018  & 0.958 & 0.913 & 1.000 & 0.938 & 0.968 & 0.111 \\
NB15+WUSTL-IIoT-2021        & ACI         & 0.333 & 0.000 & 0.000 & 0.000 & 0.000 & 0.667 \\
NB15+WUSTL-IIoT-2021        & CICIDS2017  & 1.000 & 1.000 & 1.000 & 1.000 & 1.000 & 0.000 \\
NB15+WUSTL-IIoT-2021        & CICIDS2018  & 0.958 & 0.913 & 1.000 & 0.938 & 0.968 & 0.111 \\
NB15+WUSTL-IIoT-2021        & TON         & 0.479 & 0.227 & 0.889 & 0.250 & 0.390 & 0.615 \\
TON+ACI                & CICIDS2017  & 1.000 & 1.000 & 1.000 & 1.000 & 1.000 & 0.000 \\
TON+ACI                & CICIDS2018  & 0.896 & 0.802 & 1.000 & 0.844 & 0.915 & 0.238 \\
TON+ACI                & NB15        & 0.812 & 0.584 & 0.781 & 1.000 & 0.877 & 0.000 \\
TON+WUSTL-IIoT-2021         & ACI         & 0.333 & 0.000 & 0.000 & 0.000 & 0.000 & 0.667 \\
TON+WUSTL-IIoT-2021         & CICIDS2017  & 1.000 & 1.000 & 1.000 & 1.000 & 1.000 & 0.000 \\
TON+WUSTL-IIoT-2021         & CICIDS2018  & 0.958 & 0.913 & 1.000 & 0.938 & 0.968 & 0.111 \\
TON+WUSTL-IIoT-2021         & NB15        & 0.854 & 0.679 & 0.821 & 1.000 & 0.901 & 0.000 \\
\bottomrule
\end{tabular}
\caption{Cross-domain P3 results for part 2 of the 44-experiment subset from the previous runs. \textbf{Source} lists the two training domains (A+B); \textbf{Target} is the held-out zero-shot domain.}
\label{tab:cross_domain_previous_44_part2}
\end{table*}

\subsection{Cross-Domain Benchmarking and Latent Alignment via t-SNE}
\label{subsec:latent_tsne}

\begin{figure*}[t]
\centering
\setlength{\tabcolsep}{6pt}
\renewcommand{\arraystretch}{1.1}
\begin{tabular}{ccc}
\includegraphics[width=.30\textwidth]{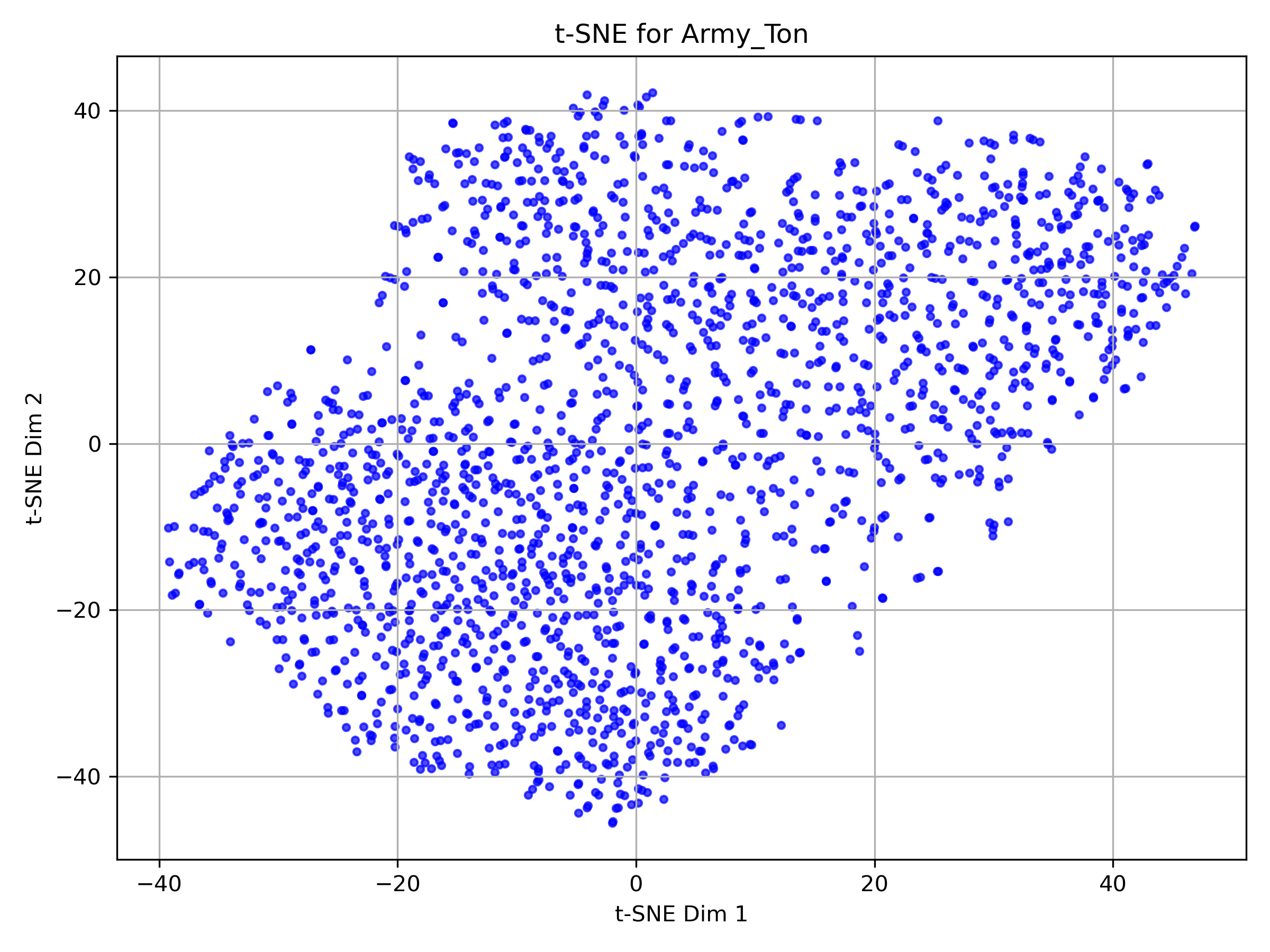} &
\includegraphics[width=.30\textwidth]{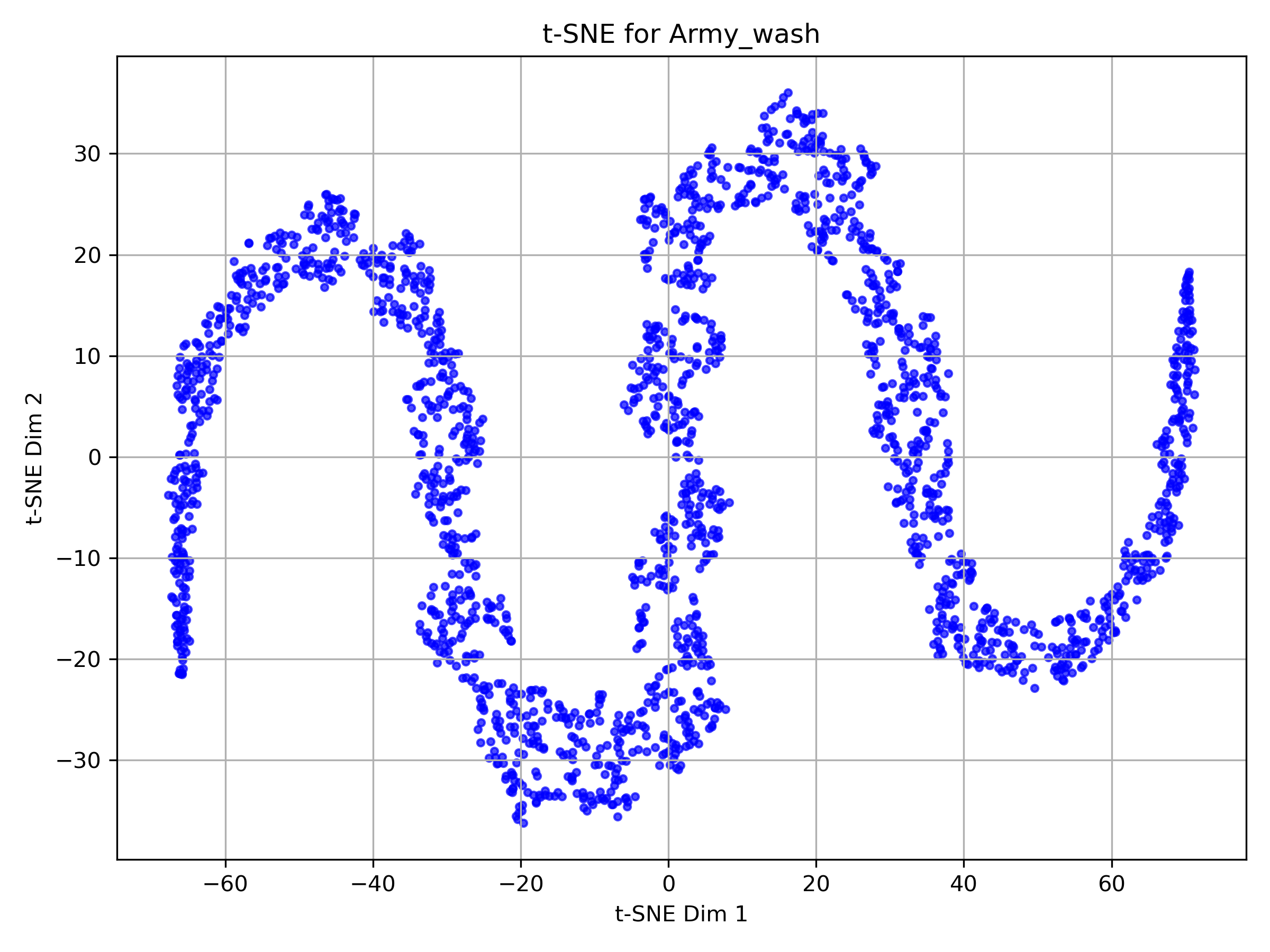} &
\includegraphics[width=.30\textwidth]{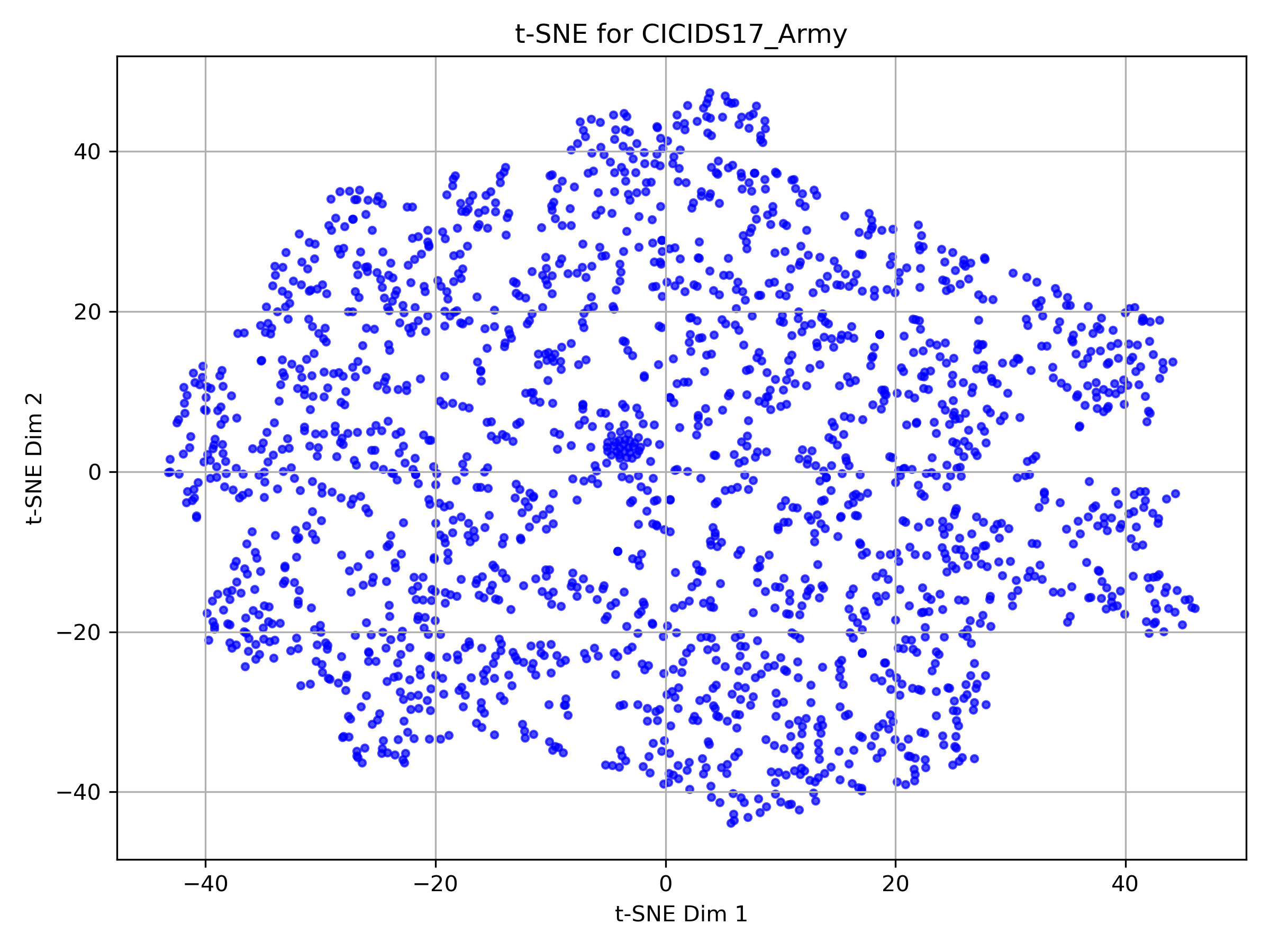} \\
{\scriptsize ACI+TON} & {\scriptsize ACI+WUSTL-IIoT-2021} & {\scriptsize CICIDS2017+ACI} \\[6pt]

\includegraphics[width=.30\textwidth]{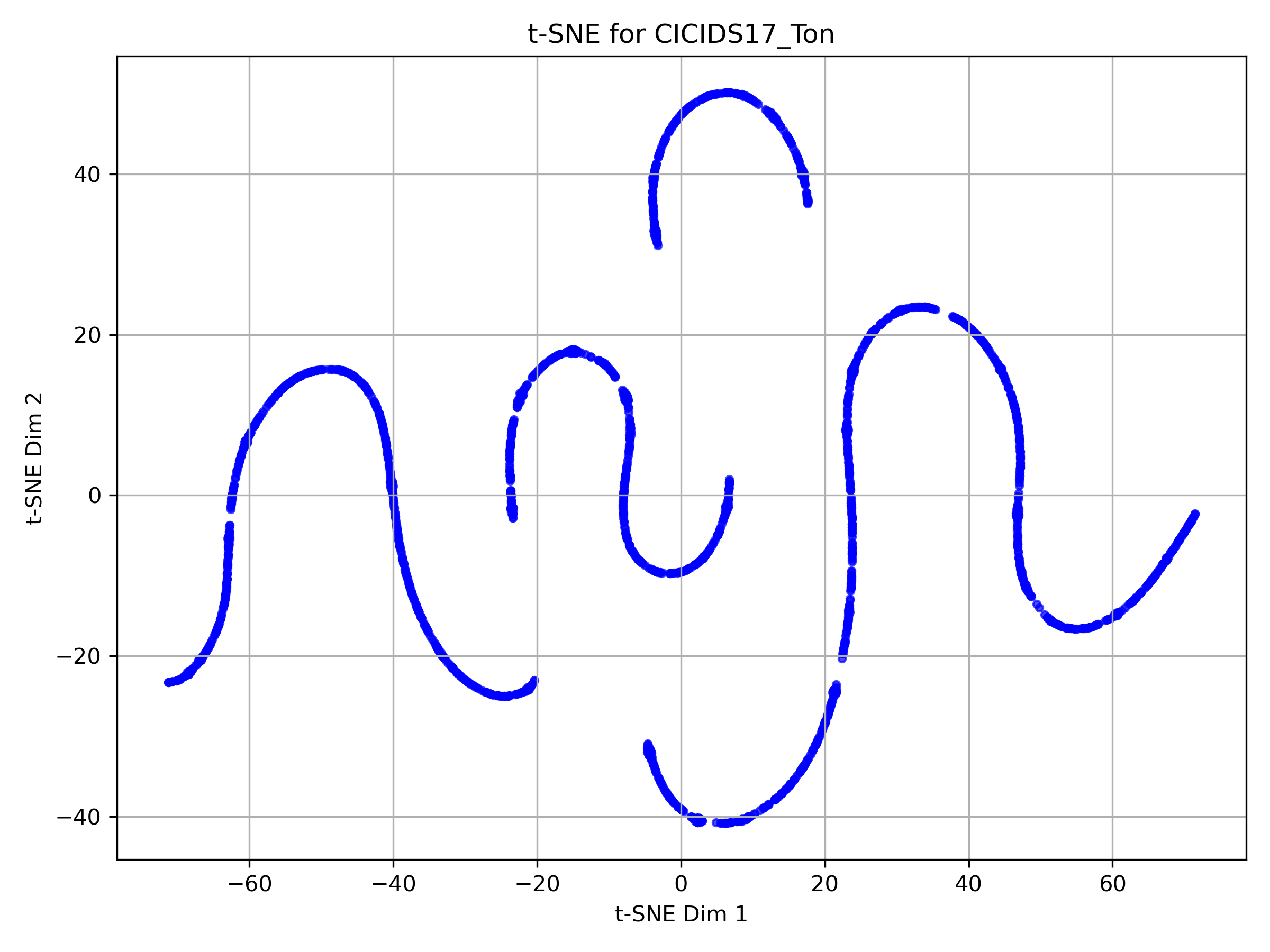} &
\includegraphics[width=.30\textwidth]{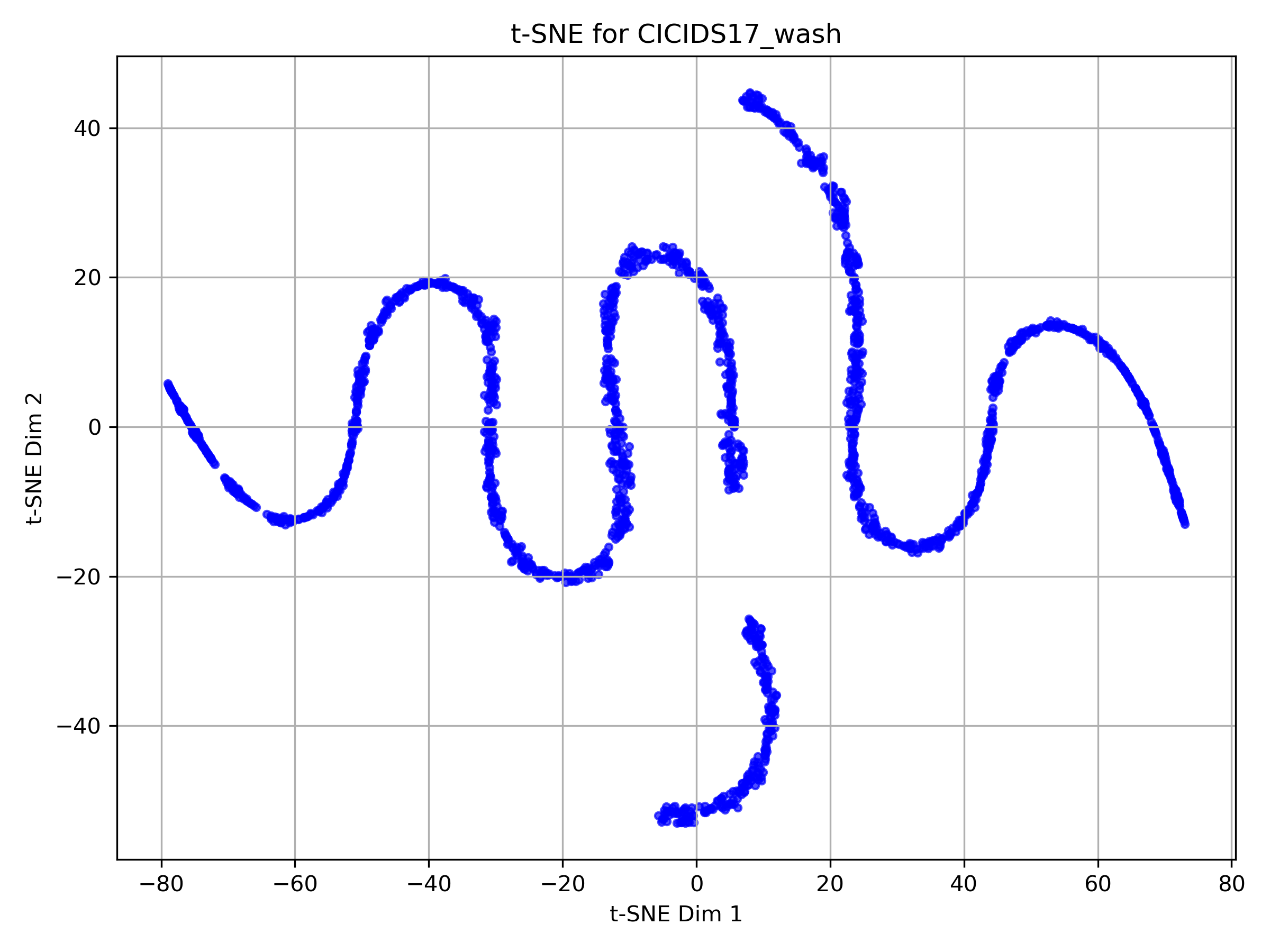} &
\includegraphics[width=.30\textwidth]{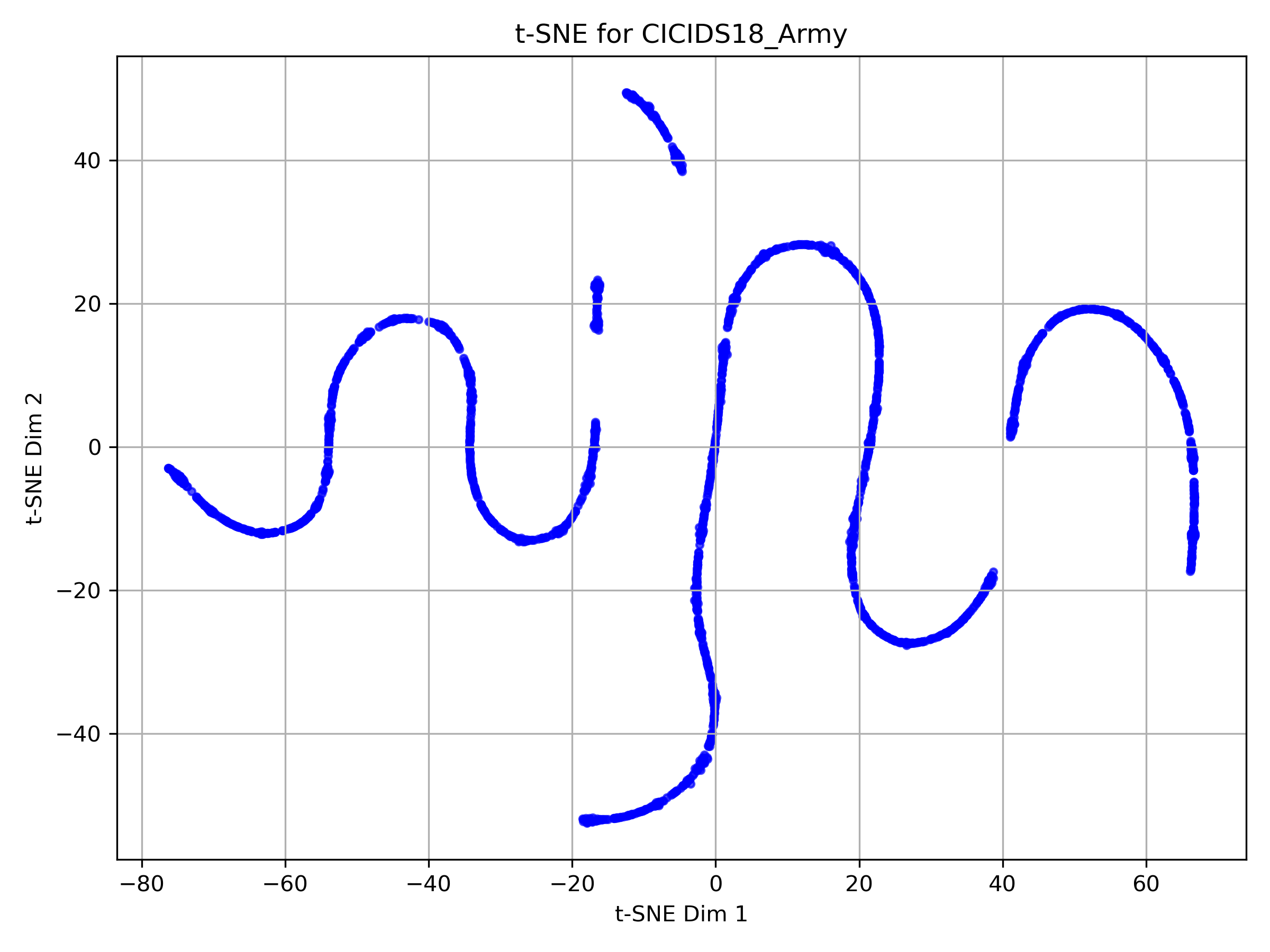} \\
{\scriptsize CICIDS2017+TON} & {\scriptsize CICIDS2017+WUSTL-IIoT-2021} & {\scriptsize CICIDS2018+ACI} \\[6pt]

\includegraphics[width=.30\textwidth]{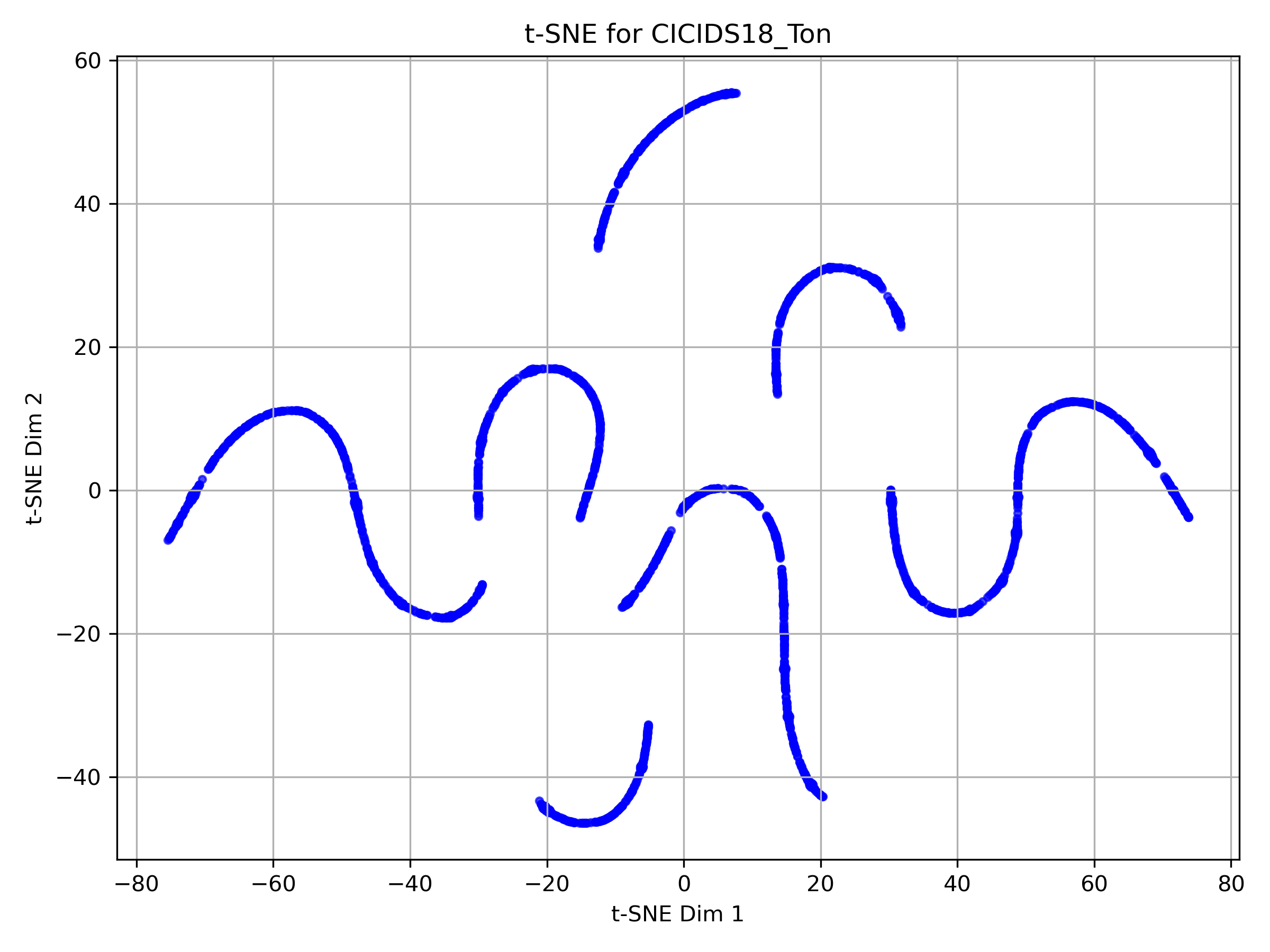} &
\includegraphics[width=.30\textwidth]{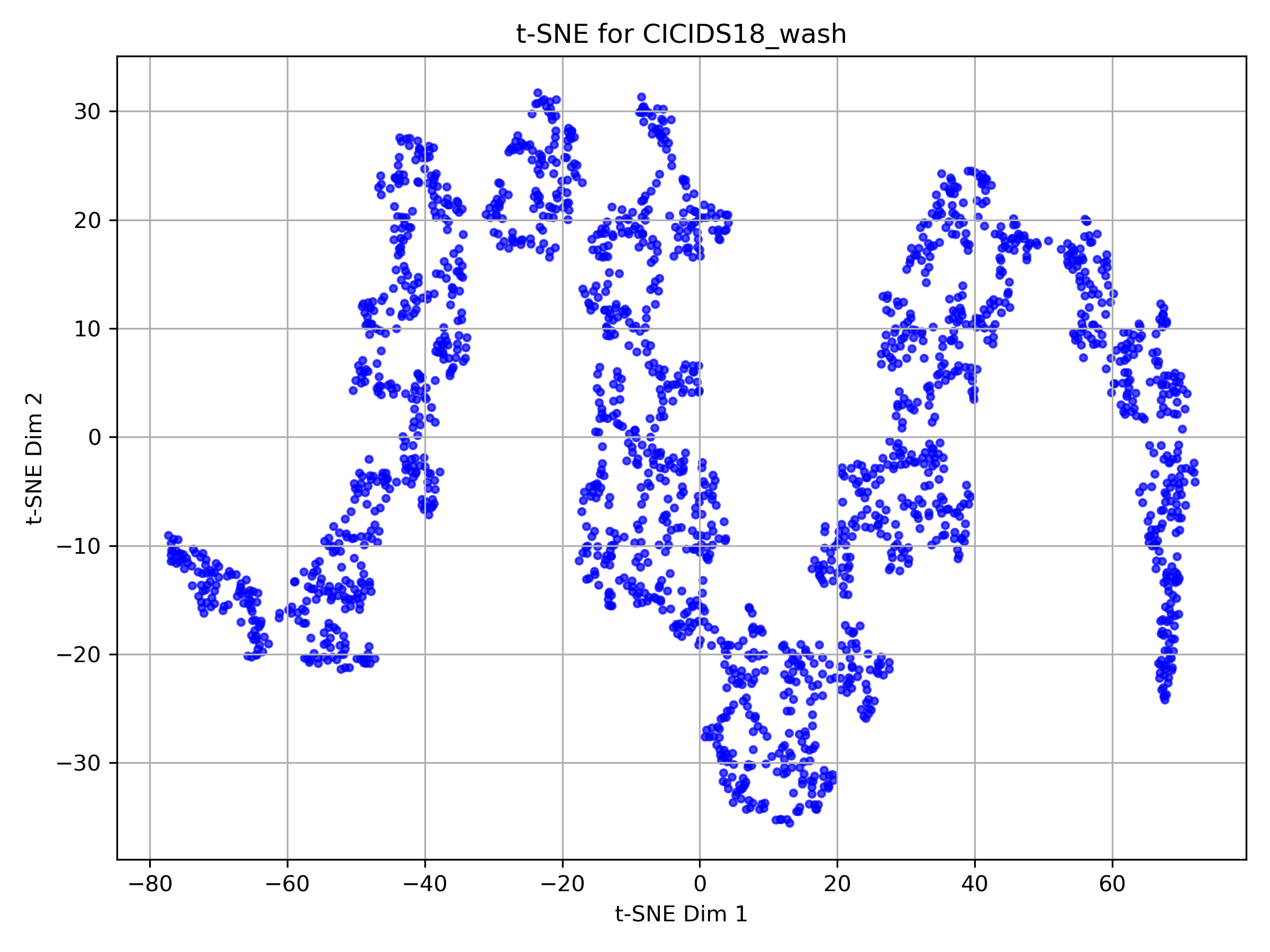} &
\includegraphics[width=.30\textwidth]{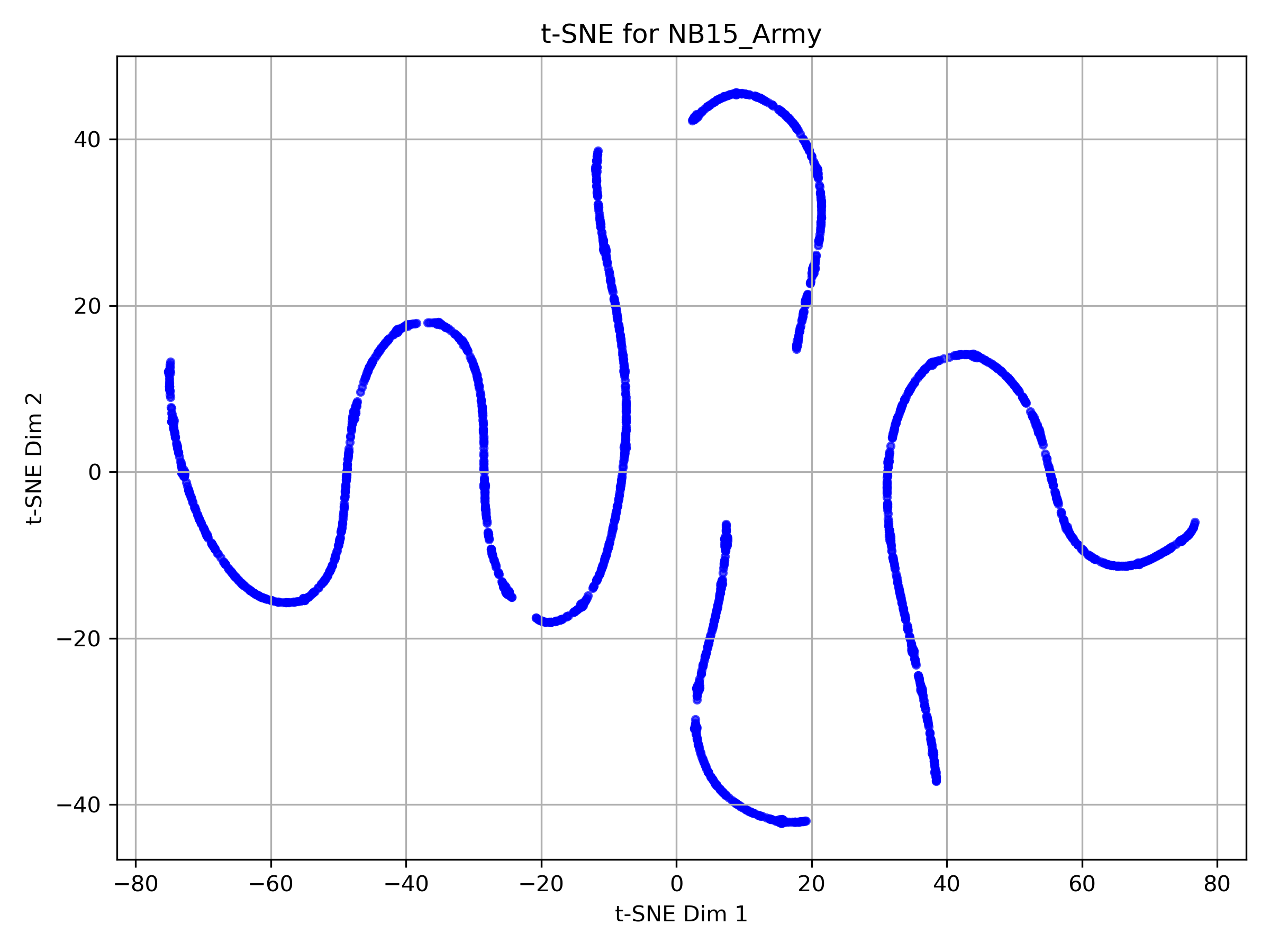} \\
{\scriptsize CICIDS2018+TON} & {\scriptsize CICIDS2018+WUSTL-IIoT-2021} & {\scriptsize NB15+ACI} \\[6pt]

\includegraphics[width=.30\textwidth]{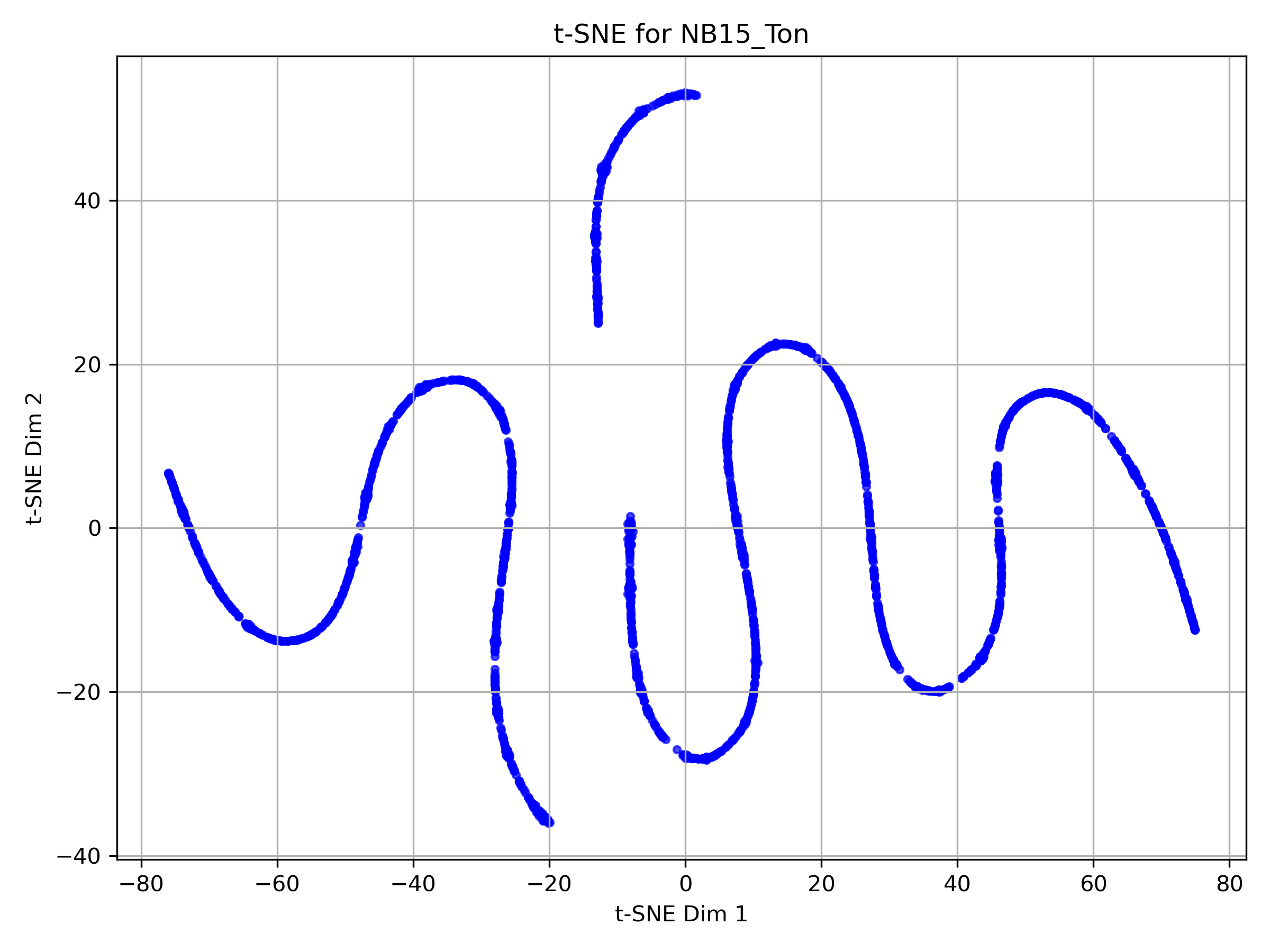} &
\includegraphics[width=.30\textwidth]{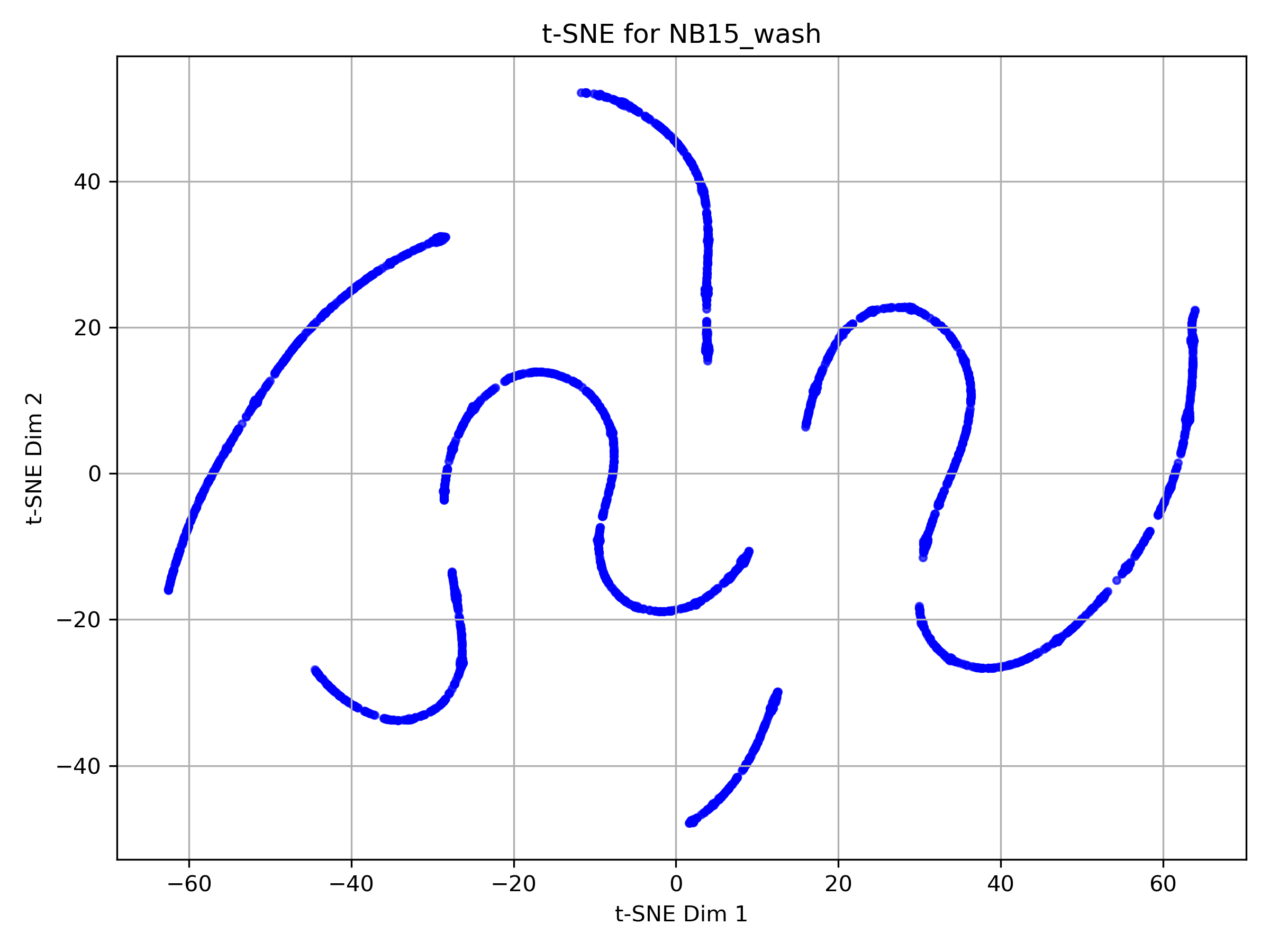} &
\includegraphics[width=.30\textwidth]{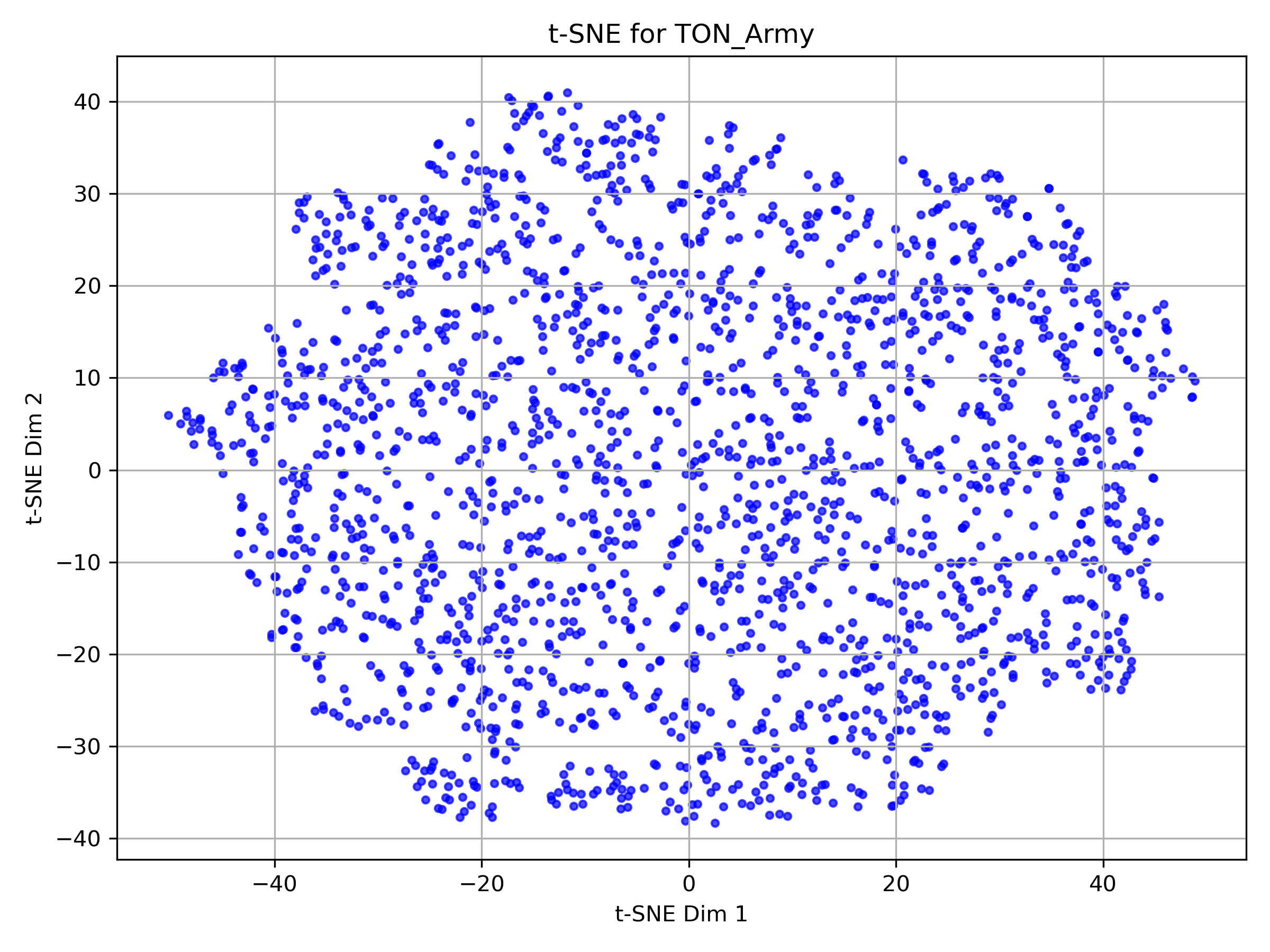} \\
{\scriptsize NB15+TON} & {\scriptsize NB15+WUSTL-IIoT-2021} & {\scriptsize TON+ACI} \\[6pt]

\includegraphics[width=.30\textwidth]{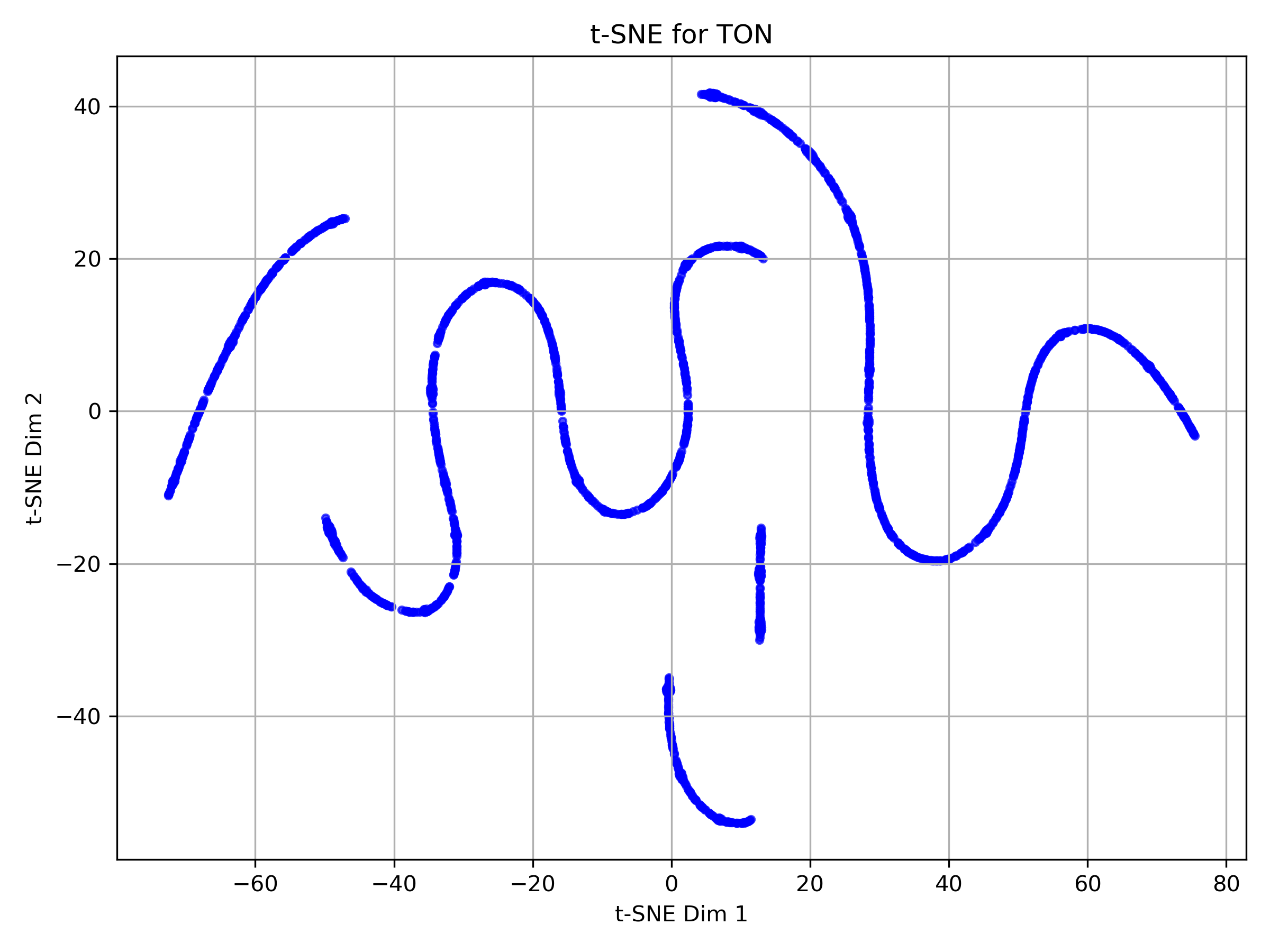} & & \\
{\scriptsize TON+WUSTL-IIoT-2021} & & 
\end{tabular}
\caption{Phase 2 latent space visualization after contrastive and adversarial alignment. Each panel shows the t-SNE projection of the aligned VAE latent representations for one source domain pair.}
\label{fig:tsne_phase2}
\end{figure*}

We expanded the evaluation to six datasets: CICIDS2017, CICIDS2018,
UNSW--NB15, TON--IoT, ACI--IoT--2023, and WUSTL-IIoT-2021/WUSTL-IIoT.
These datasets form 44 source--auxiliary--target combinations, where two
domains are used during training and alignment and the remaining domain is
held out for zero-shot evaluation. The cross-domain results are reported in
Tables~\ref{tab:cross_domain_previous_44_part1}
and~\ref{tab:cross_domain_previous_44_part2}. Figure~\ref{fig:tsne_phase2}
complements these numerical results by showing the phase--2 latent geometry,
i.e., the aligned $z$ space after VAE encoding with contrastive and
adversarial objectives, for thirteen representative source-domain pairs.

Several regularities emerge from the latent visualizations. First, source
pairs involving CICIDS2017, CICIDS2018, NB15, or WUSTL-IIoT-2021 often form smooth
low-dimensional manifolds, suggesting that the adversarial and contrastive
objectives reduce visible domain separation in the learned latent space.
Second, when TON participates as a source, the embedding often becomes highly
smooth and nearly one-dimensional. Although this indicates strong geometric
regularization, it may also suggest reduced latent diversity, which can limit
coverage of unseen target-domain modes. Third, source pairs involving ACI
sometimes retain more scattered or multi-lobed structures, indicating that
complete cross-domain mixing is more difficult when military-IoT traffic is
involved. These visual patterns suggest that latent compactness alone is not
sufficient; the aligned representation must also preserve enough variability
to support unseen target domains.

The numerical results in Tables~\ref{tab:cross_domain_previous_44_part1}
and~\ref{tab:cross_domain_previous_44_part2} show that the proposed method
performs strongly for several CICIDS2017, CICIDS2018, and NB15 target
settings. When CICIDS2017 is used as the target, source pairs such as
ACI+TON, ACI+WUSTL-IIoT-2021, and CICIDS2018+ACI achieve perfect or near-perfect
accuracy, MCC, recall, precision, and F1. Similar behavior is observed when
CICIDS2018 is the target, especially for source pairs involving CICIDS2017,
NB15, TON, or WUSTL-IIoT-2021.

By contrast, the results also reveal systematic weaknesses in some target
domains. Transfers to TON--IoT often produce low precision and F1 despite high
recall, indicating that the model detects many anomalous samples but also
generates many false positives. This behavior appears across several TON
target cases, especially when the source pairs include ACI, CICIDS2017,
CICIDS2018, NB15, or WUSTL-IIoT-2021. A similar failure mode appears when ACI is
the unseen target for several source pairs, where the model collapses toward
poor recall and low MCC. These results indicate that strict zero-shot transfer
remains challenging when the held-out target contains traffic modes or attack
distributions that are poorly covered by the aligned source domains.

Overall, the cross-domain benchmark shows that the proposed adversarial and
contrastive LSTM--VAE can learn useful transferable representations across
several heterogeneous IoT and network-security datasets, especially when the
target domain shares latent structure with the source and auxiliary domains.
At the same time, the weaker TON and ACI target cases highlight an important
limitation of strict zero-shot anomaly detection: alignment between the
available source domains does not guarantee sufficient latent support for every
unseen deployment. This motivates the DACAD comparison in the next subsection
and suggests that future work should incorporate uncertainty-aware rejection,
broader source-domain coverage, or lightweight adaptor-only calibration when a
small amount of target data is available.

\subsection{Comparison with DACAD}
\label{sec:dacad_comparison}

DACAD~\cite{DACAD} is the closest conceptual baseline to our work because it also addresses anomaly detection in multivariate time series under domain shift using contrastive domain adaptation. Its main objective is to learn transferable representations by aligning source and target domains through contrastive learning, thereby improving anomaly detection when the deployment distribution differs from the training distribution. This makes DACAD a strong reference point for evaluating whether the proposed Contrastive Adversarially-Adaptive LSTM-VAE provides additional value beyond contrastive domain adaptation alone.

Despite this conceptual similarity, our framework differs from DACAD in four important ways. First, DACAD is primarily designed around domain-adaptive multivariate time-series anomaly detection, whereas our framework is designed for strict zero-shot IoT traffic anomaly detection. In our final inference phase, no target-domain samples, labels, target loss, or target-side fine-tuning are used. The model performs detection only through reconstruction error and latent deviation from the source-learned normal-behavior manifold. Second, DACAD focuses on contrastive domain adaptation, while our method combines contrastive latent structuring with adversarial domain alignment through a gradient reversal mechanism. This dual objective simultaneously encourages semantic separation between dissimilar temporal patterns and domain invariance across heterogeneous IoT traffic sources. Third, our model explicitly supports variable-length flow sequences through the LSTM-VAE formulation, which is important for realistic IoT traffic where flows differ in duration, sampling density, and attack timescale. Fourth, our destination-centric segmentation strategy groups traffic by receiver endpoint, preserving device-level communication context instead of relying only on generic temporal windows.

To compare against DACAD, we use the same cross-domain evaluation setting adopted in the previous subsection. Multiple datasets are used as source/adversarial training domains, while the held-out target domain is never exposed during training. The comparison therefore evaluates generalization under domain shift rather than in-domain detection. This is particularly important because IoT anomaly detection systems are often deployed in environments whose device composition, protocols, traffic intensity, and attack distribution differ from those observed during training.

The comparison shows a mixed outcome rather than a uniform advantage for either method. The proposed framework performs particularly well when CICIDS2017 or CICIDS2018 is the held-out target, often reaching near-perfect accuracy in these transfer settings. In contrast, DACAD achieves stronger accuracy in several NB15, TON-IoT, and ACI-IoT target configurations. This indicates that the proposed reconstruction-based adversarial LSTM-VAE is effective when the unseen target shares temporal and flow-level structure with the aligned source domains, whereas DACAD can be more competitive when contrastive domain adaptation provides better direct representation transfer.

These results also clarify the role of the proposed design choices. Our method
emphasizes strict no-target-loss inference, variable-length sequence
reconstruction, adversarial source-domain alignment, and destination-centric
flow modeling. This design is especially effective when the held-out target
shares temporal and flow-level structure with the aligned source domains, as
seen in several CICIDS2017 and CICIDS2018 target settings. DACAD, by contrast,
is a strong contrastive domain-adaptation baseline and can outperform the
proposed method in several NB15, TON--IoT, and ACI--IoT target configurations,
where direct contrastive representation transfer appears to handle the domain
shift more effectively. Therefore, the comparison should be interpreted as
complementary rather than as a one-sided improvement: the proposed framework
offers a deployment-oriented strict zero-shot detector for IoT traffic anomaly
detection, while DACAD remains highly competitive in difficult cross-domain
accuracy comparisons.

\begin{table*}[t]
\centering
\small
\setlength{\tabcolsep}{12pt} 
\renewcommand{\arraystretch}{1.05}
\begin{tabular}{llcc}
\toprule
\textbf{Source} & \textbf{Target} & \textbf{Proposed ACC} & \textbf{DACAD ACC} \\
\midrule
ACI+TON                & CICIDS2017  & \textbf{1.000} & 0.519 \\
ACI+TON                & CICIDS2018  & \textbf{0.833} & 0.415 \\
ACI+TON                & NB15        & \textbf{0.812} & 0.649 \\
ACI+WUSTL-IIoT-2021         & CICIDS2017  & \textbf{1.000} & 0.656 \\
ACI+WUSTL-IIoT-2021         & CICIDS2018  & \textbf{0.854} & 0.738 \\
ACI+WUSTL-IIoT-2021         & NB15        & 0.833 & \textbf{0.985} \\
ACI+WUSTL-IIoT-2021         & TON         & 0.479 & \textbf{0.966} \\
CICIDS2017+ACI         & CICIDS2018  & \textbf{0.875} & 0.506 \\
CICIDS2017+ACI         & NB15        & 0.812 & \textbf{0.895} \\
CICIDS2017+ACI         & TON         & 0.479 & \textbf{0.800} \\
CICIDS2017+TON         & ACI         & 0.333 & \textbf{0.485} \\
CICIDS2017+TON         & CICIDS2018  & \textbf{0.958} & 0.551 \\
CICIDS2017+TON         & NB15        & 0.854 & \textbf{0.988} \\
CICIDS2017+WUSTL-IIoT-2021  & ACI         & 0.333 & \textbf{0.512} \\
CICIDS2017+WUSTL-IIoT-2021  & CICIDS2018  & \textbf{0.958} & 0.559 \\
CICIDS2017+WUSTL-IIoT-2021  & NB15        & 0.833 & \textbf{0.984} \\
CICIDS2017+WUSTL-IIoT-2021  & TON         & 0.479 & \textbf{0.979} \\
CICIDS2018+ACI         & CICIDS2017  & \textbf{1.000} & 0.626 \\
CICIDS2018+ACI         & NB15        & 0.833 & \textbf{0.989} \\
CICIDS2018+ACI         & TON         & 0.479 & \textbf{0.563} \\
CICIDS2018+TON         & ACI         & 0.333 & \textbf{0.492} \\
CICIDS2018+TON         & CICIDS2017  & \textbf{1.000} & 0.600 \\
\bottomrule
\end{tabular}
\caption{Accuracy comparison between the proposed method and DACAD for part 1 of the 44-experiment P3 subset. The higher accuracy in each row is shown in bold.}
\label{tab:previous_vs_dacad_acc_44_part1}
\end{table*}

\begin{table*}[t]
\centering
\small
\setlength{\tabcolsep}{12pt} 
\renewcommand{\arraystretch}{1.05}
\begin{tabular}{llcc}
\toprule
\textbf{Source} & \textbf{Target} & \textbf{Proposed ACC} & \textbf{DACAD ACC} \\
\midrule
CICIDS2018+TON         & NB15        & 0.812 & \textbf{0.891} \\
CICIDS2018+WUSTL-IIoT-2021  & ACI         & 0.333 & \textbf{0.418} \\
CICIDS2018+WUSTL-IIoT-2021  & CICIDS2017  & \textbf{1.000} & 0.691 \\
CICIDS2018+WUSTL-IIoT-2021  & NB15        & 0.854 & \textbf{0.904} \\
CICIDS2018+WUSTL-IIoT-2021  & TON         & 0.479 & \textbf{0.963} \\
NB15+ACI               & CICIDS2017  & \textbf{1.000} & 0.484 \\
NB15+ACI               & CICIDS2018  & \textbf{0.938} & 0.545 \\
NB15+ACI               & TON         & 0.479 & \textbf{0.603} \\
NB15+TON               & ACI         & 0.333 & \textbf{0.564} \\
NB15+TON               & CICIDS2017  & \textbf{1.000} & 0.647 \\
NB15+TON               & CICIDS2018  & \textbf{0.958} & 0.594 \\
NB15+WUSTL-IIoT-2021        & ACI         & 0.333 & \textbf{0.418} \\
NB15+WUSTL-IIoT-2021        & CICIDS2017  & \textbf{1.000} & 0.620 \\
NB15+WUSTL-IIoT-2021        & CICIDS2018  & \textbf{0.958} & 0.551 \\
NB15+WUSTL-IIoT-2021        & TON         & 0.479 & \textbf{0.984} \\
TON+ACI                & CICIDS2017  & \textbf{1.000} & 0.389 \\
TON+ACI                & CICIDS2018  & \textbf{0.896} & 0.410 \\
TON+ACI                & NB15        & \textbf{0.812} & 0.632 \\
TON+WUSTL-IIoT-2021         & ACI         & 0.333 & \textbf{0.449} \\
TON+WUSTL-IIoT-2021         & CICIDS2017  & \textbf{1.000} & 0.680 \\
TON+WUSTL-IIoT-2021         & CICIDS2018  & \textbf{0.958} & 0.755 \\
TON+WUSTL-IIoT-2021         & NB15        & 0.854 & \textbf{0.940} \\
\bottomrule
\end{tabular}
\caption{Accuracy comparison between the proposed method and DACAD for part 2 of the 44-experiment P3 subset. The higher accuracy in each row is shown in bold.}
\label{tab:previous_vs_dacad_acc_44_part2}
\end{table*}

\section{Conclusion}

This paper introduced a contrastive adversarially adaptive LSTM--VAE framework
for zero-shot anomaly detection in multivariate IoT network traffic. By
combining sequence-based variational reconstruction with adversarial
source-domain alignment and cosine-based contrastive latent structuring, the
framework addresses cross-domain shifts without requiring target-side labels,
target-domain loss computation, or target-side fine-tuning. The integration of
lightweight domain adaptors and destination-centric traffic segmentation further
allows the model to accommodate variable-length flows while preserving
device-level communication context.

Extensive evaluation across six diverse IoT and network-security benchmarks
spanning 44 transfer scenarios demonstrates that the proposed framework achieves
strong zero-shot generalization in several structurally compatible
cross-domain settings, particularly for CICIDS2017 and CICIDS2018 targets. At
the same time, the results expose systematic degradation in highly shifted
smart-home and military-IoT target domains when the aligned source domains do
not provide sufficient latent coverage. In particular, TON--IoT target cases
often produce high recall but low precision, indicating elevated false-positive
rates, while several ACI--IoT target cases show poor recall and low MCC under
strict zero-shot transfer.

Comparison with DACAD shows a complementary performance pattern rather than a
uniform advantage for either method. The proposed reconstruction-driven
LSTM--VAE is highly effective when temporal and flow-level structures transfer
across domains, whereas direct contrastive domain adaptation can be more
competitive under severe distribution shift. These findings highlight both the
promise and the limitations of strict zero-shot anomaly detection for
heterogeneous IoT deployments.

\section*{CRediT authorship contribution statement}
Mahshid Rezakhani: Conceptualization, Methodology, Software, Validation,
Formal analysis, Investigation, Data curation, Writing -- original draft,
Visualization. Tolunay Seyfi: Conceptualization, Methodology, Software,
Validation, Formal analysis, Investigation, Data curation, Writing -- original
draft, Writing -- review and editing, Visualization. Fatemeh Afghah:
Conceptualization, Supervision, Project administration, Funding acquisition,
Writing -- review and editing.

\section*{Declaration of competing interest}
The authors declare that they have no known competing financial interests or
personal relationships that could have appeared to influence the work reported
in this paper.

\section*{Data availability}
The datasets analyzed in this study are publicly available from their respective
repositories. Processed data and implementation details are available from the
corresponding author upon reasonable request.

\section*{Acknowledgements}
This material is based upon work supported by the National Science Foundation
under Grant Numbers CNS-2318726 and CNS-2232048.

\nocite{*}
\bibliographystyle{IEEEtran}
\bibliography{cas-refs}

\end{document}